\documentclass[runningheads]{llncs}

\usepackage{eccv}
\usepackage{amssymb}

\usepackage{eccvabbrv}
\usepackage{dblfloatfix}

\usepackage{graphicx}
\usepackage{booktabs}
\usepackage{multirow}
\usepackage{threeparttable}
\usepackage{pgfplots}
\usepackage{adjustbox}
\pgfplotsset{compat=1.18}

\usepackage[accsupp]{axessibility}  

\usepackage{hyperref}

\usepackage{orcidlink}

\begin{document}

\title{Identity-Consistent Video Generation under Large Facial-Angle Variations}

\titlerunning{Multi-view reference for identity consistency}

\author{
Bin Hu\inst{1,2}\textsuperscript{*} \and
Zipeng Qi\inst{2}\textsuperscript{*} \and
Guoxi Huang\inst{3} \and
Zunnan Xu\inst{1} \and
Ruicheng Zhang\inst{1} \and
Chongjie Ye\inst{4} \and
Jun Zhou\inst{1} \and
Xiu Li\inst{1}\textsuperscript{\dag} \and
Jingdong Wang\inst{2}\textsuperscript{\dag}\textsuperscript{\ddag}
}

\authorrunning{B.~Hu et al.}

\institute{Tsinghua University \and
Baidu Inc., China \and University of Bristol \and The Chinese University of Hong Kong, Shenzhen}

\maketitle

\begingroup
\renewcommand\thefootnote{}
\footnotetext{\textsuperscript{*} Equal contribution.}
\footnotetext{\textsuperscript{\dag} Corresponding authors.}
\footnotetext{\textsuperscript{\ddag} Project leader.}
\endgroup

\begin{figure*}
    \centering
    \includegraphics[width=1.0\linewidth]{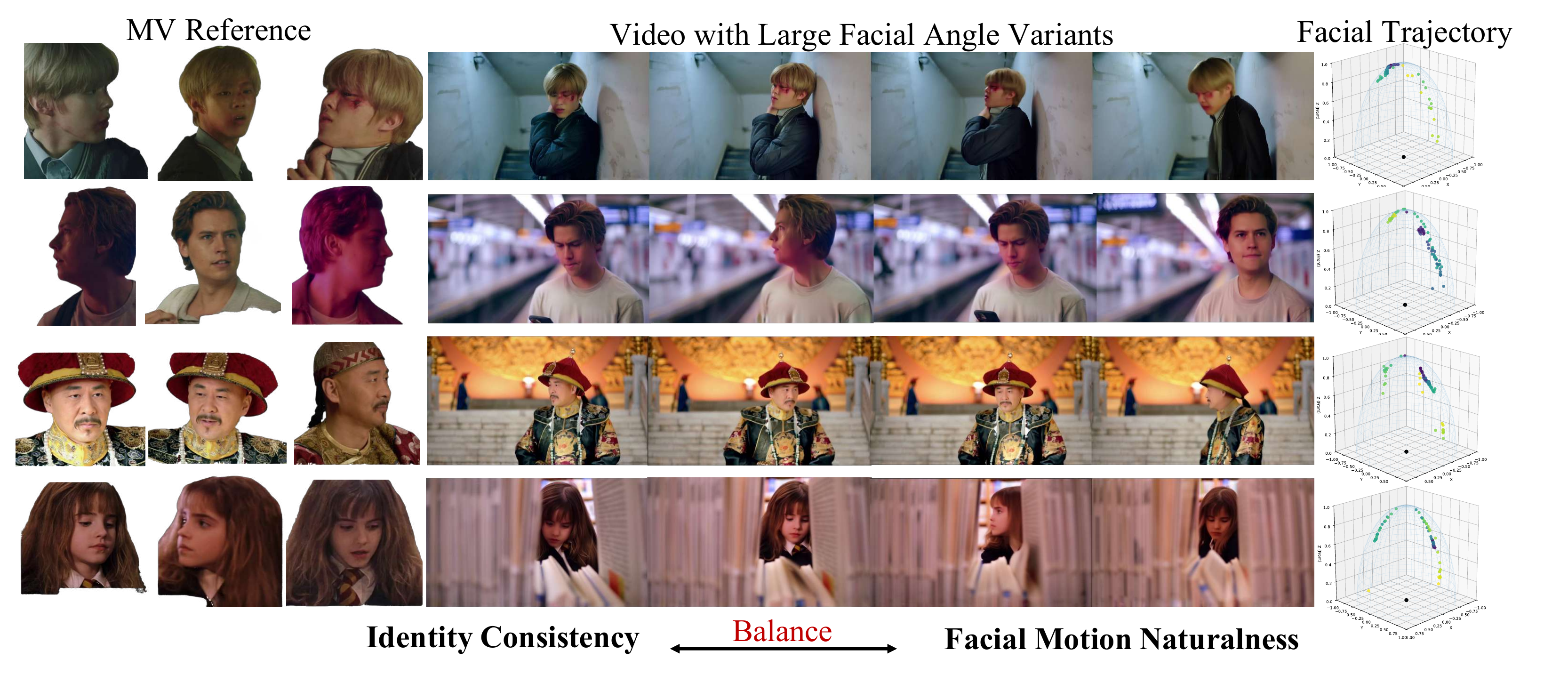}
    \caption{In this work, we leverage multi-view references to improve identity consistency under large facial angle variations, while maintaining motion naturalness (see enlarged view for trajectory details).}
    \label{fig:placeholde}
\end{figure*}

\begin{abstract}
Single-view reference-to-video methods often struggle to preserve identity consistency under large facial-angle variations. This limitation naturally motivates the incorporation of multi-view facial references. However, simply introducing additional reference images exacerbates the \textit{copy-paste} problem, particularly the \textbf{\textit{view-dependent copy-paste}} artifact, which reduces facial motion naturalness. Although cross-paired data can alleviate this issue, collecting such data is costly. To balance the consistency and naturalness, we propose $\mathrm{Mv}^2\mathrm{ID}$, a multi-view conditioned framework under in-paired supervision. We introduce a region-masking training strategy to prevent shortcut learning and extract essential identity features by encouraging the model to aggregate complementary identity cues across views. In addition, we design a reference decoupled-RoPE mechanism that assigns distinct positional encoding to video and conditioning tokens for better modeling of their heterogeneous properties. Furthermore, we construct a large-scale dataset with diverse facial-angle variations and propose dedicated evaluation metrics for identity consistency and motion naturalness. Extensive experiments demonstrate that our method significantly improves identity consistency while maintaining motion naturalness, outperforming existing approaches trained with cross-paired data.
\keywords{Video Generation \and Identity Consistency \and Copy-paste}
\end{abstract}

\section{Introduction}
\label{sec:intro}
Diffusion-based generative models have become the dominant paradigm for image~\cite{ ldm2022,improved_ddpm2021,edm2022, qi2024layered, qi2023difftalker}, video~\cite{video_diffusion_2022,imagen_video2022,make_a_video2022,phenaki2022,cogvideo2022,lumiere2024,wan2025,hunyuanvideo15_2025,opensora_plan2024,opensora2024,huang2026consistentid}, and 3D generation~\cite{dreamfusion2022,magic3d2022,prolificdreamer2023,zero1232023,zhao2025gshoi}.
Among them, reference-to-video approaches enable controllable video synthesis conditioned on visual references.
While prior efforts have improved prompt alignment~\cite{wan2025,hunyuanvideo15_2025,opensora_plan2024}
and reference-based appearance consistency~\cite{jiang2024videobooth,Wei_2024_CVPR,Wu_2023_ICCV,Li_2025_ICCV,Liu_2025_ICCV,Zhang2025Kaleido,hong2025audio,xu2025hunyuanportrait}, identity consistency under large facial-angle variations remains largely underexplored.

This challenge becomes more pronounced in identity-known scenarios, where the target identity (e.g., a specific individual) is predefined and must remain consistent across diverse viewpoints. Compared with generic appearance consistency, this setting imposes significantly stricter requirements on identity consistency and motion coherence.

\begin{figure}
    \centering
    \includegraphics[width=1.0\linewidth]{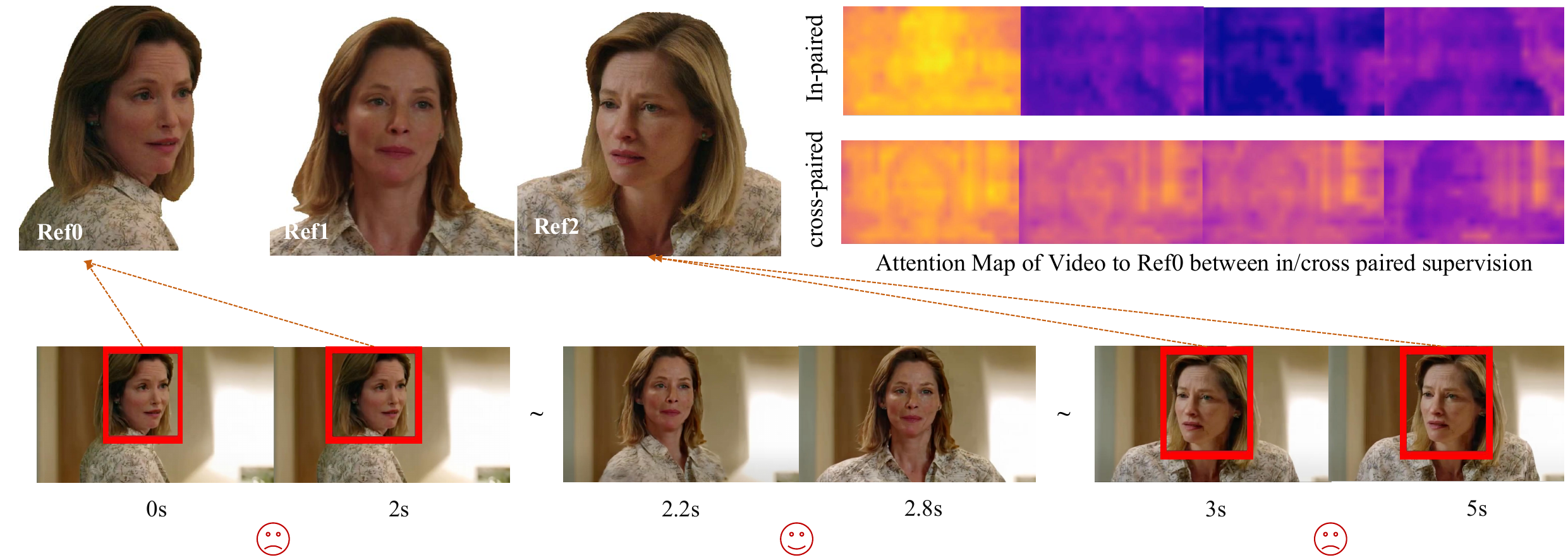}
    \caption{\textit{Copy-paste}: Within certain time intervals, the generated video collapses to a specific reference image, resulting in reduced motion diversity(more detailed analysis can be found in the \textbf{Observation} in Sec.~\ref{sec:prerequisites}).}
    \label{fig:problem}
\end{figure}

From an information-theoretic perspective, a single-view reference provides insufficient information to preserve identity consistency across diverse facial angles. As a result, single-view based methods~\cite{Li_2025_ICCV,jiang2024videobooth} are inherently limited under large facial-angle variations. This fundamental limitation naturally motivates the incorporation of multi-view facial references. However, naively introducing additional reference images can exacerbate the copy-paste problem, especially for the \textit{view-dependent copy-paste} effect\footnote{Unless otherwise stated, we denote \textit{copy-paste} as viewing-angle \textit{copy-paste}}(see the right of Figure~\ref{fig:problem}), where generated faces overly concentrate around conditioning facial angles, leading to reduced motion diversity (unnaturalness). Notably, this issue has received limited systematic study.

Although cross-paired data~\cite{yuan2025opensvnexus,phantom_data2025} can mitigate copy-paste artifacts, acquiring high-quality cross-paired supervision is substantially more expensive, especially under large-angle variations. In practice, less than 5\% of available data meets the requirements for reliable multi-view conditioning. Therefore, a key challenge is how to enhance identity consistency while preserving motion naturalness using low-cost in-paired data.

To address this challenge, we propose Mv$^{2}$ID, a multi-view--conditioned video generation framework. Our core idea is to leverage complementary identity cues from multiple views while preventing shortcut learning for mitigating the \textit{copy-paste}. We adopt a token-level integration scheme that appends facial conditioning tokens to video tokens and enables cross-token interaction via self-attention. Furthermore, we introduce a region masking training strategy that enforces identity information aggregation across views rather than reliance on a single perspective. This can discourage view-specific shortcuts and extract more essential identity features. In addition, we propose a reference-decoupled RoPE mechanism that assigns distinct positional encoding to video and conditioning tokens, enabling more effective modeling of their heterogeneous temporal and structural characteristics. Furthermore, we construct a large-scale human-centric video dataset with multi-view facial conditions, comprising 22K videos and a test set covering over 30 identities. We also introduce dedicated evaluation metrics for identity consistency and motion naturalness. Extensive experiments demonstrate that Mv$^{2}$ID significantly improves identity consistency while preserving realistic motion, outperforming existing methods trained with cross-paired data.

Our contributions are summarized as follows:
\begin{itemize}
    \item We identify the limitations of single-view conditioning for preserving identity under large facial-angle variations and introduce a dedicated multi-view conditioning framework for human-centric video generation.

    \item To enhance consistency while mitigating the \textit{copy-paste} issue under low-cost in-paired supervision, we design two novel components: a region masking training strategy and a reference-decoupled RoPE mechanism.

    \item We construct a large-scale multi-view video \textit{dataset} and propose dedicated \textit{evaluation metrics}. Extensive experiments demonstrate the effectiveness of our approach under in-paired supervision.
\end{itemize}

\section{Related Work}
\label{sec:related_work}

\noindent\textbf{Video generation models.}
Modern video generators largely follow a latent-space paradigm: raw videos are compressed by video VAEs/autoencoders, and large-scale generative pre-training is conducted with Transformer-based backbones (typically diffusion transformers).
Recent open-source foundation models, including Wan~\cite{wan2025}, HunyuanVideo 1.5~\cite{hunyuanvideo15_2025}, and Open-Sora~\cite{opensora2024, opensora_plan2024} show that stronger spatiotemporal compression, scalable training, and efficient attention design consistently improve visual fidelity and motion coherence for both text-to-video and image-to-video generation~\cite{wan2025,hunyuanvideo15_2025,opensora2024}.

\noindent\textbf{Reference-to-video.}
Reference-to-video (R2V) methods condition generation on text prompts and one or more reference images to preserve subject identity and appearance over time.
Recent representative methods include Phantom~\cite{Liu_2025_ICCV}, MAGREF~\cite{Deng2025MAGREF}, and HuMo~\cite{chen2025humo}, which substantially improve subject consistency under standard settings.
Existing approaches are mainly grouped into two lines:
(1) \emph{feature extraction and injection}, which compresses references into global embeddings (e.g., CLIP/ID features) and injects them via cross-attention, adapters, LoRA, as in VideoBooth, DreamVideo, PersonalVideo, and Tune-A-Video~\cite{jiang2024videobooth,Wei_2024_CVPR,Li_2025_ICCV,Wu_2023_ICCV};
(2) \emph{latent/token fusion}, which encodes references into the same latent/token space as video tokens and performs joint fusion, as in VACE~\cite{vace2025}.
However, explicit handling of multi-view conditioning under large viewpoint changes remains limited, and large-angle cross-view identity consistency is usually not the primary optimization target~\cite{Liu_2025_ICCV,Deng2025MAGREF,chen2025humo}.

\noindent\textbf{Copy-paste / shortcut learning.}
Under \emph{in-paired} supervision, models may exploit shot-level correlations and exhibit shortcut behaviors such as attention dominance on a single reference view and viewpoint-trajectory collapse.
Recent studies analyze such failures and propose mitigation strategies, including cross-paired data construction, masking/augmentation, and identity-dynamics decoupling~\cite{shortcut_mitigation_diffusion2023,Zhang2025Kaleido,Liu_2025_ICCV,Li_2025_ICCV,Wei_2024_CVPR,yuan2025opensvnexus,phantom_data2025}.
OpenS2V-Nexus~\cite{yuan2025opensvnexus}    explicitly operationalize \emph{copy-paste}-related failure using a dedicated naturalness-oriented protocol (NaturalScore), which evaluates whether identity consistency is achieved without sacrificing motion plausibility and viewpoint transition smoothness.

\section{Prerequisites}
\label{sec:prerequisites}
In this section, we detail the notions and pre-knowledge of diffusion models, as well as represent an important pre-observation.

\textbf{Notions}. Suppose a video generation model $\mathcal{M}$ generates a clean video sample $z_0$ conditioned on a text prompt $c$ and a set of multi-view identity image conditions, denoted as $\{I_1, \ldots, I_m\}$.

\textbf{Flow-based diffusion models}. Diffusion models generate data by transforming Gaussian noise into samples through a reverse-time process~\cite{ddpm2020,improved_ddpm2021,score_sde2021}. In the standard discrete formulation, a neural network predicts noise at each timestep to gradually denoise $Z_T \sim \mathcal{N}(0,\mathbf{I})$ toward a clean sample $z_0$.

Beyond this discrete view, diffusion models admit a continuous-time formulation~\cite{score_sde2021}. In particular, the reverse stochastic differential equation (SDE) is associated with a deterministic \emph{probability flow ODE}, which defines a continuous trajectory from noise to data:
\begin{equation}
\frac{\mathrm{d} z}{\mathrm{d}t} = v(z, t),
\end{equation}
where $v(z,t)$ denotes a time-dependent velocity field. Flow-based diffusion models~\cite{flow_matching2023,rectified_flow2022}, such as flow matching and rectified flows, directly parameterize the velocity field $v_\theta(z,t)$ to transport samples along deterministic trajectories from the source distribution to the data distribution~\cite{flow_matching2023,rectified_flow2022}. 
Instead of predicting noise, the model learns the conditional flow governing sample evolution over time, interpreting generation as trajectory integration in latent space and providing a flexible framework for conditioning.

\textbf{Observation.}
Naively using multiple reference views as conditioning (e.g., three images) can improve consistency (as shown in Figure~\ref{fig:ablation_vis}; additional results are provided in the supplementary material.) but often leads to a \textit{copy-paste} artifact(shown in Fig.~\ref{fig:problem}), especially under in-paired supervision.

\textit{Copy-paste. The generated video may remain locked to a conditioning facial angle for a period of time (e.g., 0–2s in Figure~\ref{fig:problem}), followed by an abrupt switch to another reference angle, resulting in unnatural motion.}

Attention analysis (shown in top right part of Figure~\ref{fig:problem}) further reveals that, under in-paired training, the model concentrates excessively on a particular reference view during certain time segments, causing view-dependent copy-paste. In contrast, cross-paired supervision produces more balanced cross-view attention, which aligns with smoother and more natural visual transitions. However, high-quality cross-paired data is difficult to obtain, particularly under large-angle variations. Therefore, a central challenge is how to improve identity consistency while preserving motion naturalness under low-cost in-paired supervision.

\section{Methodology}
\label{sec:method}
\subsection{Large-angle Dataset Construction}
We construct the dataset through a three-stage pipeline that preserves scale while explicitly enriching large-pose coverage. First, we perform coarse filtering~\cite{talebi2018nima,imagereward2023} on raw videos using face detection~\cite{retinaface2019} to ensure the presence of visible faces while removing low-quality content and segments without humans. Second, we segment videos into coherent clips, discard short clips, and apply face detection and tracking to retain samples containing a single human with long and stable face tracks. Third, we estimate 3D head pose (yaw, pitch, roll) for the retained clips to mine large-angle examples (e.g., yaw $>45^\circ$ or pitch $>45^\circ$), and generate or refine captions using Qwen2.5-VL~\cite{qwen2025qwen25technicalreport}. More implementation details and the complete pipeline are provided in the supplementary material.

\subsection{Multi-view Reference Injection}
\begin{figure*}
    \centering
    \includegraphics[width=0.98\linewidth]{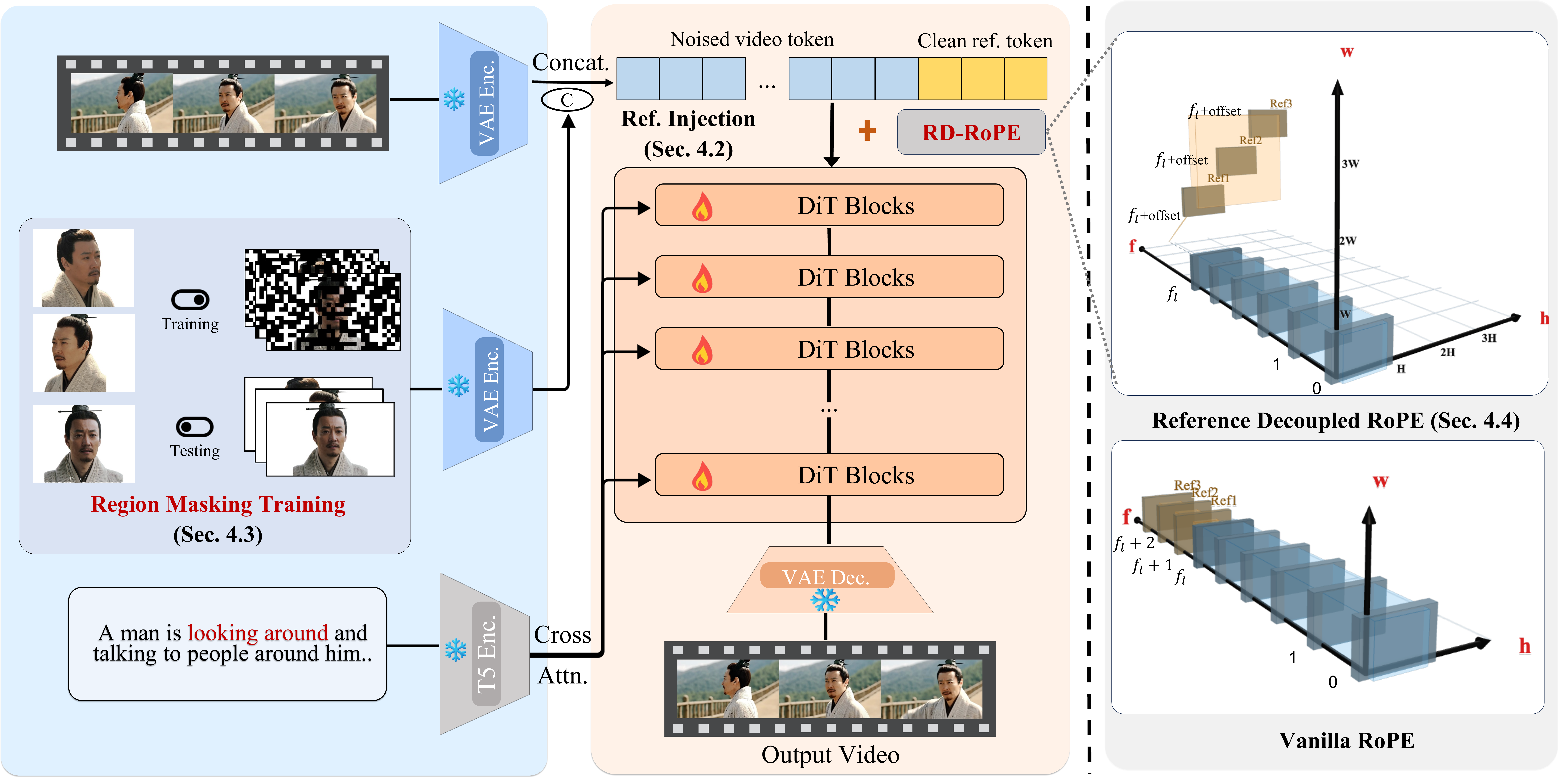}
    \caption{The overview of the training pipeline. We design the region masking training(Sec. 4.3) and reference-decoupled RoPE(Sec. 4.4). For inference, we use the clean condition images. }
    \label{fig:overview}
\end{figure*}

We adopt a widely used conditioning injection strategy that concatenating the noisy video tokens with the clean reference tokens along the sequence dimension(as shown in Figure~\ref{fig:overview}). Formally, the input sequence is constructed as
\begin{equation}
\mathbf{X} = [\{\mathbf{X}_i^t\}_{i=1}^{m+1}, \mathbf{R}_1, \mathbf{R}_2, \cdots, \mathbf{R}_n],
\label{eq:cond_concat}
\end{equation}
where $\{\mathbf{X}_i^t\}_{i=1}^{m+1}$ denotes the noised video latent at time $t$. The clean latent is obtained by encoding the video $z_0$ with a VAE(e.g., WanVAE~\cite{wan2025}). The $\{\mathbf{R}_i\}_{i=1}^{n}$ represents $n$ reference latents, split extracted by the same VAE encoder. We input the concatenated sequence into the DiT-based model and make information interaction between the visual tokens via self-attention mechanism. This design keeps the backbone architecture intact and facilitates efficient and stable training by minimizing architectural changes. In addition, we also adopt a text encoder (e.g., T5~\cite{t5_2020}) to extract textual embeddings from the input prompt $c$, which are interacted into the model via cross attention.  

\textbf{Reason to \textit{copy-paste}.}
The \textit{copy-paste} artifact stems from shortcut learning during training, similar to appearance-copy phenomena~\cite{shortcut_mitigation_diffusion2023,Liu2024DiffShortcut}. To minimize reconstruction loss, the model may adopt a low-cost strategy by directly reusing appearance cues from conditioning references instead of learning view-consistent identity representations. This results in over-reliance on specific views, ignoring other view, as reflected by the attention concentration in Fig.~\ref{fig:problem}.  Cross-paired supervision mitigates this behavior by making direct copying less feasible and promoting balanced cross-view aggregation. Our key insight is therefore to restrict shortcut pathways and discourage trivial copying under in-paired supervision.

\subsection{Region Masking Training}
In this section, we start from a straightforward strategy termed View Masking (VM). For each video latent token, we prevent the model from attending to the reference tokens with the closest facial angle (see the supplementary material for details). Formally, the attention map is computed as follows:
\begin{equation}
A = \mathrm{Attention}(Q = X, K = X, V = X).
\end{equation}
We then set $\mathbf{A}_{i,j}=0$ to mask the attention between tokens from the $i_{th}$ video latent and tokens from the $j_{th}$ reference latent. The $j_{th}$ reference latent corresponds to the reference image whose facial angle is closest to the first of four video frames associated with the $i_{th}$ video latent. However, this naive and straightway has two limitations. First, it cannot fully eliminate shortcut information flow, since indirect interactions may still occur through other tokens via the self-attention mechanism. Second, it reduces the available reference information, which contradicts our goal of enhancing multi-view information aggregation.

Inspired by Masked Autoencoders (MAE~\cite{mae2021}), which learn essential representations by reconstructing masked regions from partially observed inputs, we propose a \textbf{Region Masking (RM)} strategy. 
Unlike VM, which blocks view-level attention, RM randomly masks spatial regions of all reference images at a fixed ratio (e.g., 60\%), as illustrated in Fig.~\ref{fig:overview}. Specifically, we first apply a binary mask map $M$ to each reference image $I_i$ in the pixel space. 
The mask map has the same spatial size as the image, where a fixed proportion of entries are randomly set to 0 and the remaining entries are set to 1. The masked reference latent is then obtained as
\begin{equation}
R_i^m = \mathrm{VAE}(I_i \odot M),
\end{equation}
where $R_i^m$ denotes the latent representation extracted from the masked reference image. This design serves two purposes. First, it prevents direct copying entire information from arbitrary condition image, addressing the first limitation of VM. Second, by partially removing information from every conditioning image, it forces the model to aggregate complementary cues across multi-view to generate the target frame, alleviating the second limitation of VM. Consequently, RM encourages the model to learn view-invariant identity representations, improving identity consistency while mitigating \textit{copy-paste} artifacts caused by over-reliance on specific conditioning images. More comparison between VM and RM and experiments with different ratio settings are provided in supplementary material.

\subsection{Reference-Decoupled RoPE}
To better model both intra-video dependencies and video–reference interactions, we introduce a different RoPE mechanism for video and reference. We first start from the vanilla unified 3D RoPE design. In the unified setting, the position of each token is defined by a shared 3D coordinate system $\{f,h,w\}$, where $f$ denotes the index of belonged frame, and $(h,w)$ denote the spatial coordinates within that frame. The spatial coordinates $(h,w)$ are aligned across video and reference tokens. Meanwhile, reference tokens are assigned temporal indices that follow the video frames in the sequence dimension:
\begin{equation}
f_i = f_l + i,
\end{equation}
where $f_l$ denotes the index of the last video frame, and $f_i$ denotes the assigned temporal index of the $i_{th}$ reference image. However, this unified design has two drawbacks. Video and reference tokens share the same coordinate system, obscuring the separation between temporal modeling and cross-view interactions. Moreover, assigning artificial temporal indices to reference images introduces inappropriate inductive bias.

Thus, we propose \textbf{RD-RoPE} (as shown in Figure~\ref{fig:overview}), a fully decoupled coordinate design compared with the vanilla 3D RoPE. Our key insight is that \emph{multi-view reference images should be temporally equivalent}. Following previous methods\cite{wan2025,hunyuanvideo15_2025}, we assign a temporal offset to reference tokens to differentiate them from video frames:
\begin{equation}
    f_i = f_l + \mathrm{offset}, \quad i = 1,2,\dots,n,
\end{equation}
However, this design still treats reference images as temporally subsequent frames of the video, which may confuse the model and introduce imbalance in positional encoding across temporal and spatial dimensions (see supplementary materials for analysis). To address this issue, we instead allocate distinct spatial coordinate regions for each reference token:
\begin{equation}
\begin{aligned}
    h_i &= h_v + i \cdot H, \\
    w_i &= w_v + i \cdot W,
\end{aligned}
\quad i = 1,2,\dots,n,
\end{equation}
where $H$ and $W$ denote the spatial grid dimensions, $n$ denotes the number of reference tokens, and $(h_v, w_v)$ represent the spatial coordinates tokens under the vanilla RoPE coordinate system. This design makes all reference images temporally equivalent while distinguishing them through spatial coordinates. More importantly, it fully decouples reference tokens from the video’s coordinate system, enabling more effective modeling of cross-view interactions.


\section{Empirical Analysis}
\label{sec:exp}

\subsection{Experimental Setup}
\label{sec:exp_setup}
In this section, we first show our experiment details and then present the comparison and ablation results.

\textbf{Implementation Details.} For the main experiments, we build our method on Wan-2.1-T2V-14B\cite{wan2025} and use three reference images by default. All training videos are resampled to 16 fps and resized to a spatial resolution of $480 \times 832$. 
For ablation studies, we adopt a smaller base model with a similar architecture, Wan-2.2-TI2V-5B\cite{wan2025}, to improve experimental efficiency.

\textbf{Testing data.} We select 30 identities, each paired with 5 text prompts and 3 reference images, resulting in 150 test samples. Since our work focuses on large facial-angle variations, the prompts mainly describe facial motions (e.g., looking back, talking, and head movements). For identity evaluation, we additionally use 10 corresponding multi-view images per identity to provide more comprehensive viewpoint coverage. This test set is independently curated from a different domain disjoint from the training data.

\textbf{Evaluation Metrics.}
We evaluate all methods using six metrics covering perceptual quality, semantic alignment, human motion quality, identity consistency, and facial motion naturalness. 
For the first four aspects, we adopt AES, IQA, TVA, and HA from VBench~\cite{vbench2023}. Specifically, AES and IQA measure perceptual quality, TVA assesses text–video alignment, and HA evaluates human motion quality.

To evaluate identity consistency under large facial-angle variations, we propose \textbf{MvRC} (\textbf{M}ulti-\textbf{v}iew \textbf{R}eference \textbf{C}onsistency), which measures identity preservation across multiple viewpoints. For $i_{th}$ generated frame $z_0^i$, we crop the facial region and extract face embeddings using a feature extractor $M_f$. Let $\mathcal{R}=\{r_1,\dots,r_{10}\}$ denote the reference set. The frame-level consistency is defined as:
\begin{equation}
\mathrm{MvRC}(z^i_0)=\mathrm{avg}_{r \in \mathcal{R}}\, \mathrm{cosine}\!\left(M_f(z^i_0), M_f(r)\right),
\end{equation}
where $\mathrm{cosine}(\cdot,\cdot)$ denotes cosine similarity between face embeddings. We adopt ArcFace\cite{arcface2018} and CurricularFace~\cite{Huang_2020_CVPR} as the feature extractors and corresponding metrics denoted as MvRC-Arc and MvRC-Cur, respectively. The final video-level score is obtained by averaging the frame-level scores over sampled frames.

To evaluate facial motion naturalness (i.e., less \textit{copy-paste}, smoother dynamics), we adopt the NaturalScore protocol from OpenS2V-Nexus~\cite{yuan2025opensvnexus}. 
Following their setup, we use the same evaluation prompt and replace GPT-4o~\cite{openai2024gpt4o} with the more recent vision-language model Seed2.0-Lite~\cite{bytedance2026seed2} to score each generated video on a 1–5 scale.

\begin{table}[h]
    \centering
    \small
    \setlength{\tabcolsep}{5pt}
    \caption{
    Main comparison with prior methods under single-view (SV) and multi-view (MV) settings.
    }
    \label{tab:comparison_table}
    \begin{threeparttable}
    \begin{adjustbox}{max width=\linewidth}
    \begin{tabular}{lccccccc}
        \toprule
        \multirow{2}{*}{\textbf{Method}} 
        & \multicolumn{2}{c}{\textbf{Quality}} 
        & \textbf{Align.} 
        & \textbf{Motion} 
        & \multicolumn{2}{c}{\textbf{Identity}}
        & \textbf{Naturalness} \\
        \cmidrule(lr){2-3} 
        \cmidrule(lr){4-4} 
        \cmidrule(lr){5-5} 
        \cmidrule(lr){6-7}
        \cmidrule(lr){8-8}
        & \textbf{AES}$\uparrow$
        & \textbf{IQA}$\uparrow$
        & \textbf{TVA}$\uparrow$
        & \textbf{HA}$\uparrow$
        & \textbf{MvRC-Arc}$\uparrow$
        & \textbf{MvRC-Cur}$\uparrow$
        & \textbf{NaturalScore}$\uparrow$ \\
        \midrule
        Phantom-SV~\cite{Liu_2025_ICCV}      & 0.523 & 0.621 & 0.091 & \underline{0.977} & 0.408 & 0.365& 3.98 \\
        Phantom-MV~\cite{Liu_2025_ICCV}      & 0.532 & 0.635 & 0.091 & 0.972              & 0.487 & 0.438& 4.11 \\
        HuMo~\cite{chen2025humo}             & 0.531 & \textbf{0.649} & \underline{0.092} & \textbf{0.978} & \underline{0.493} & \underline{0.459}& \textbf{4.71} \\
        MAGREF-SV~\cite{Deng2025MAGREF}      & 0.567 & 0.602 & 0.085 & 0.947              & 0.270 & 0.223& 4.40 \\
        MAGREF-MV~\cite{Deng2025MAGREF}      & \underline{0.569} & 0.602 & 0.086 & 0.952      & 0.338 & 0.286& 4.43 \\
        \midrule
        Mv$^2$ID (Ours) & \textbf{0.568} & \underline{0.645} & \textbf{0.097} & 0.960 & \textbf{0.544} & \textbf{0.507}& \underline{4.69} \\
        \bottomrule
    \end{tabular}
    \end{adjustbox}
    \end{threeparttable}
\end{table}

\begin{figure}
    \centering
    \includegraphics[width=0.9\linewidth]{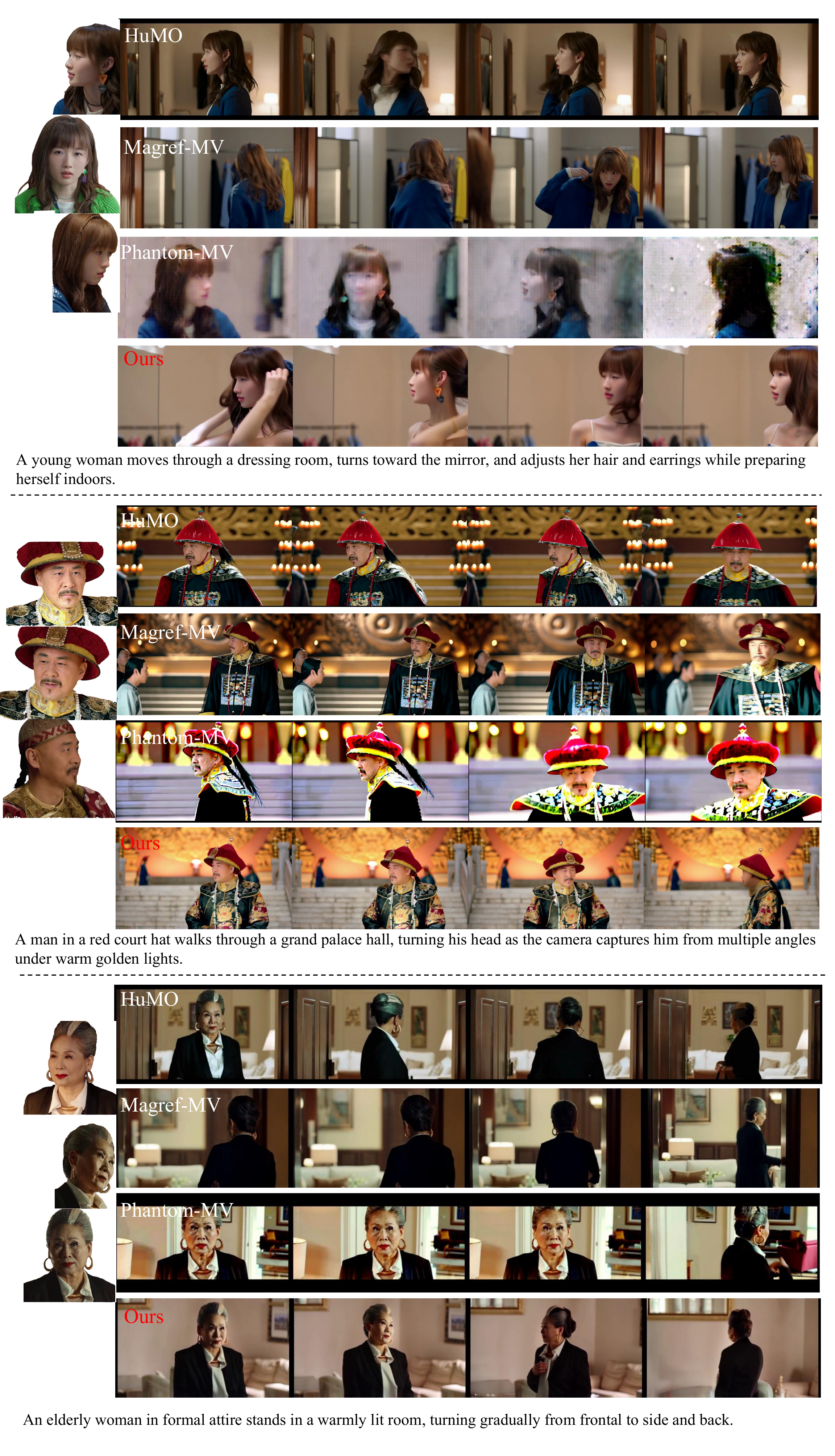}
    \caption{Qualitative comparison under large facial-angle changes. Our method,Mv$^2$ID, better balances the consistency and facial motion naturalness.}
    \label{fig:comparison_fig}
\end{figure}

\subsection{Comparison Results}
\label{sec:sota}
We compare our method with pervious methods, including HuMo, MAGREF and Phantom, which are trained with cross-paired data. For the MAGREF and Phantom, we compare on both single-view and multi-view settings\footnote{The MAGREF and Phantom support multiple image input.} (same as in \cite{song2026mvs2vmultiviewsubjectconsistentvideo}). We use the default setting for all compared methods for fairness.

\textbf{Quantitative comparison.}
Table~\ref{tab:comparison_table} shows that our method achieves better whole video quality, both in visual quality and prompt alignment. Our main concern is identity consistency. The results show that Mv$^2$ID achieves the best MvRC (\textbf{0.544}, \textbf{0.507}), outperforming all baselines, including the strongest competitor HuMo (0.493, 0.459). This confirms that our method better preserves subject identity under facial-angle changing. In addition, our method also achieves strong naturalness, with a NaturalScore of \textbf{4.69}, which is close to the best result (4.71). These more balanced results suggest that the improved consistency does not arise from \textit{copy-paste}, proving the effectiveness of our method. The advantage of the multi-view setting over the single-view methods (Phantom and MAGREF) further supports our insight that multi-view references improve identity consistency.

\begin{figure}
    \centering
    \includegraphics[width=1.0\linewidth]{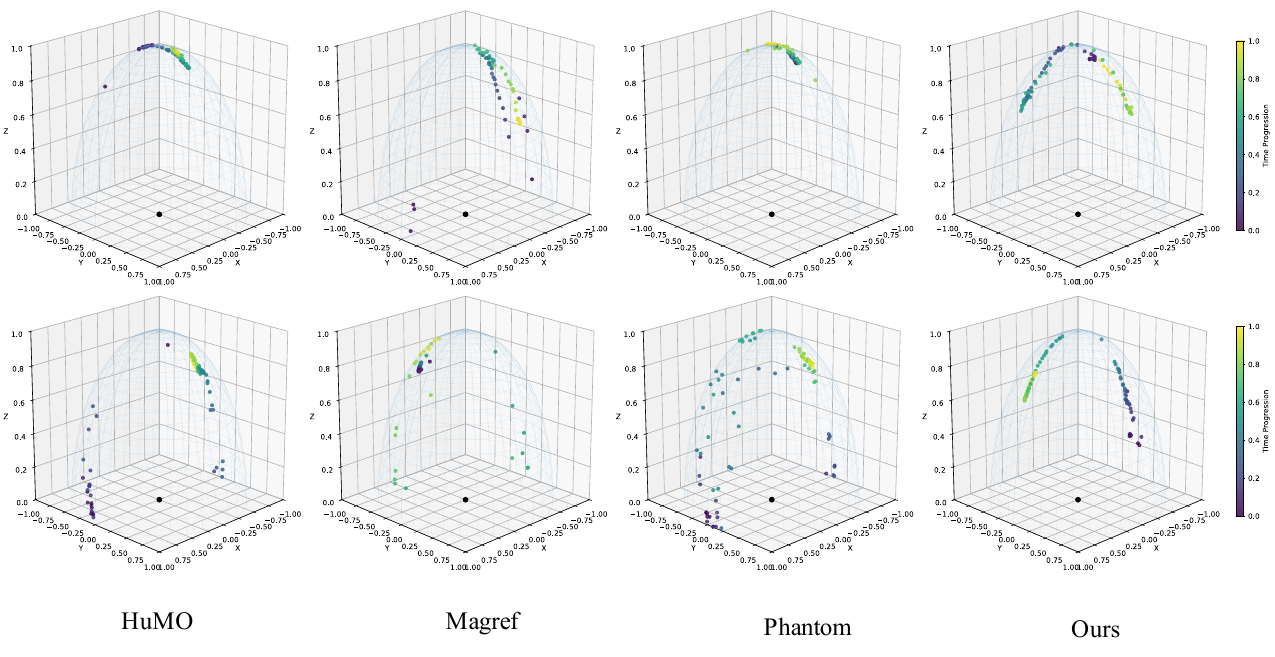}
    \caption{Facial trajectory comparison. Each dot represents the normalized 3D facial direction vector of a frame, and the black dot denotes the origin. Colors indicate the temporal progression of frames, normalized to the range [0,1] across 81 frames.}
    \label{fig:comparison_tracjectory}
\end{figure}

\textbf{Qualitative comparison.}
As shown in Fig.~\ref{fig:comparison_fig}, existing methods are more prone to identity drift. In the first example, HuMo and Phantom exhibit noticeable facial inconsistencies across frames: Phantom shows deviations in the jawline and eyebrows, while HuMo generates a frontal face that differs from the reference. MAGREF also introduces an abrupt increase in facial brightness, leading to unnatural appearance. A similar issue appears in the third example, where Phantom is strongly affected by lighting conditions in the references, and the facial motions produced by HuMo and MAGREF appear less realistic. In contrast, our method, $\mathrm{Mv}^2\mathrm{ID}$, maintains stable identity consistency (e.g., facial structure and clothing details) while preserving natural viewpoint transitions. These observations are consistent with the improvements in metrics.

\textbf{Trajectory comparison.} Figure~\ref{fig:comparison_tracjectory} shows the results of face trajectory. The dots represent the normalized 3D facial direction vector of a frame and the black dot is the origin. The color of dot indicates the temporal progression within total 81 frames. Similar colored dots clustered together indicates that the face was facing very similarly during that time, meaning it collapsed to a certain angle. If the dots are widely distributed, it indicates that the face was moving rapidly. Based on this, we can find that the results of the three comparison methods are either relatively concentrated(the first column of HuMo and the third column of Phantom) or highly dispersed(the second column of MAGREF and the third column of Phantom). In contrast, our results is more smooth and reasonable. This comparison is difficult to reflect in indicators and several visualizations.

\begin{figure}[h]
    \centering
    \includegraphics[width=0.98\linewidth]{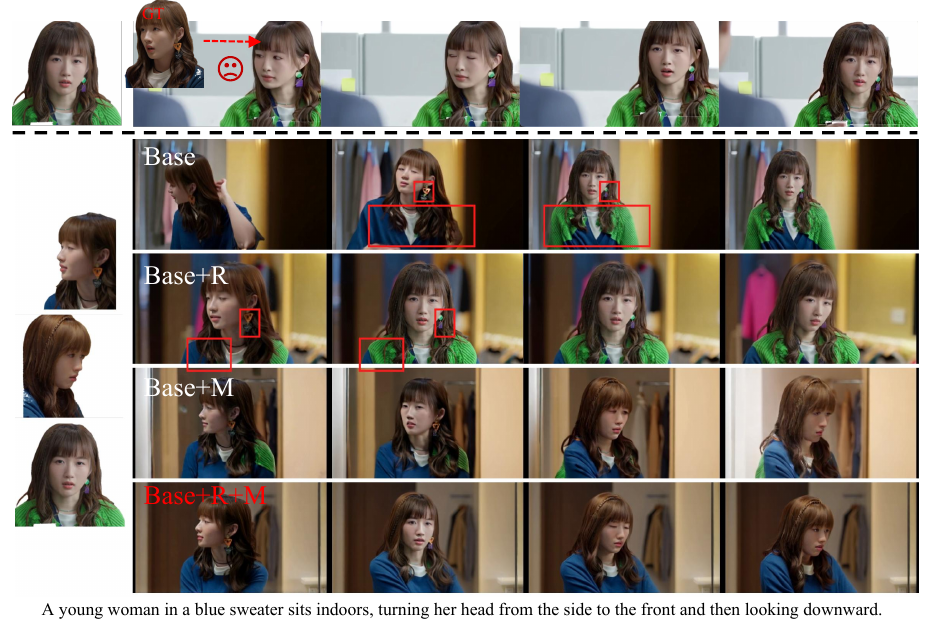}
    \caption{Single-view vs. Multi-view and qualitative comparison of proposed components. R: RD-RoPE. M : Region-Masking Training.}
    \label{fig:ablation_vis}
\end{figure}

\begin{table}[t]
    \centering
    \small
    \setlength{\tabcolsep}{5pt}
    \caption{
    Ablation study of the proposed components.
    }
    \label{tab:ablation}
    \begin{threeparttable}

    \begin{adjustbox}{max width=\linewidth}
    \begin{tabular}{lccccccc}
        \toprule
        \multirow{2}{*}{\textbf{Setting}} 
        & \multicolumn{2}{c}{\textbf{Quality}} 
        & \textbf{Align.} 
        & \textbf{Motion} 
        & \multicolumn{2}{c}{\textbf{Identity}}
        & \textbf{Naturalness} \\
        \cmidrule(lr){2-3} 
        \cmidrule(lr){4-4} 
        \cmidrule(lr){5-5} 
        \cmidrule(lr){6-7}
        \cmidrule(lr){8-8}
        & \textbf{AES}$\uparrow$
        & \textbf{IQA}$\uparrow$
        & \textbf{TVA}$\uparrow$
        & \textbf{HA}$\uparrow$
        & \textbf{MvRC-Arc}$\uparrow$
        & \textbf{MvRC-Cur}$\uparrow$
        & \textbf{NaturalScore}$\uparrow$ \\
        \midrule
        B         & 0.568 & 0.664 & 0.089 & 0.960 & \underline{0.546} & \underline{0.515} & 3.70 \\
        B + R     & \underline{0.578} & \underline{0.671} & \underline{0.091} & \underline{0.971} & \textbf{0.552} & \textbf{0.522} & \underline{4.12} \\
        B + M     & 0.569 & \textbf{0.686} & \textbf{0.093} & 0.966 & 0.537 & 0.504 & 3.93 \\
        B + R + M & \textbf{0.580} & \underline{0.671} & 0.087 & \textbf{0.975} & 0.535 & 0.502 & \textbf{4.52} \\
        \bottomrule
    \end{tabular}
    \end{adjustbox}

    \begin{tablenotes}[flushleft]
        \footnotesize
        \item \textit{B}: Base method. \textit{R}: RD-RoPE. \textit{M}: Region-Masking Training.
    \end{tablenotes}
    \end{threeparttable}
\end{table}

\subsection{Ablation Studies}
\label{sec:ablation}
In the ablation study, we evaluate RD-RoPE and region masking (RM), using direct multi-view conditioning with three reference images as the base method. 

As shown in Table~\ref{tab:ablation}, both components improve the overall quality of generated results. Region masking training and RD-RoPE improve motion naturalness by 0.52 and 0.23, respectively. When combined, the improvement further increases to \textbf{0.82}, demonstrating the effectiveness of mitigating the \textit{copy-paste} effect. Although region masking slightly reduces identity consistency(caused by the less \textit{copy-paste}), the overall level remains comparable to the baseline. Compared with other methods, the final model still achieves higher consistency while producing more natural facial motion. Notably, the benefit of region masking for identity consistency is not fully captured by quantitative metrics. We therefore present qualitative visual comparisons.

Figure~\ref{fig:ablation_vis} shows the qualitative comparison results. 
We observe that directly using multi-view references already provides a clear advantage in enhancing identity consistency, as illustrated in the top part of the figure. Moreover, certain improvements are not fully captured by quantitative metrics. For instance, both Base+M and the final model can extract consistent clothing attributes despite conflicting colors across reference images; a similar phenomenon can be observed in fine-grained details such as earrings. 
These results suggest that region masking encourages the model to focus on essential identity characteristics rather than view-specific appearance cues.

\begin{figure}
    \centering
    \includegraphics[width=1.0\linewidth]{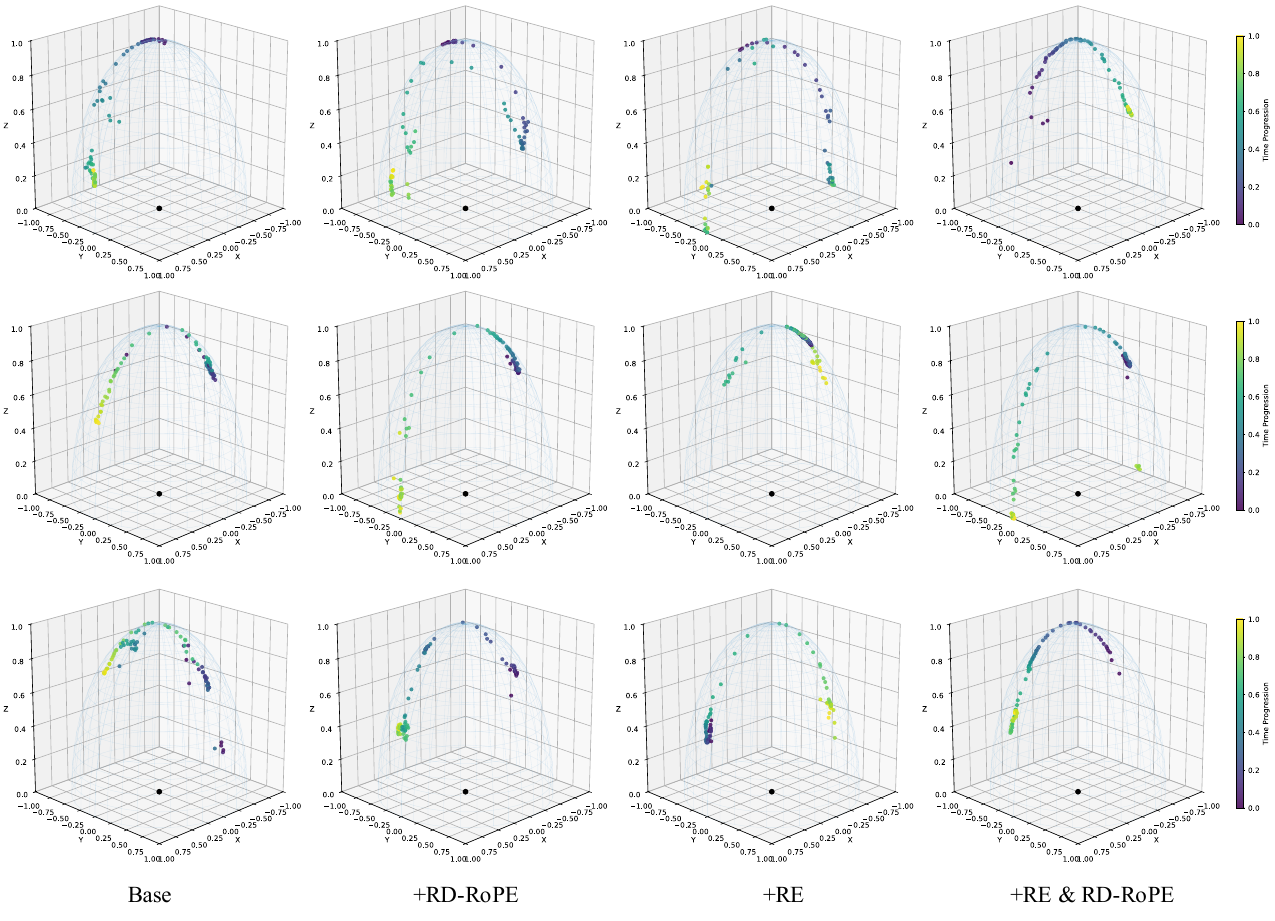}
    \caption{Facial trajectory for ablation. }
    \label{fig:ablation_tracjectory}
\end{figure}

We further visualize facial trajectories in Figure~\ref{fig:ablation_tracjectory}. Dispersed dots indicate abrupt viewpoint changes, whereas concentrated dots suggest view-dependent copy-paste. RD-RoPE alleviates collapse, RM reduces excessive dispersion, and their combination yields smoother and more natural facial rotation. 

\subsection{User Study}
\label{sec:user_study}
\begin{figure}
    \centering
    \includegraphics[width=0.78\linewidth]{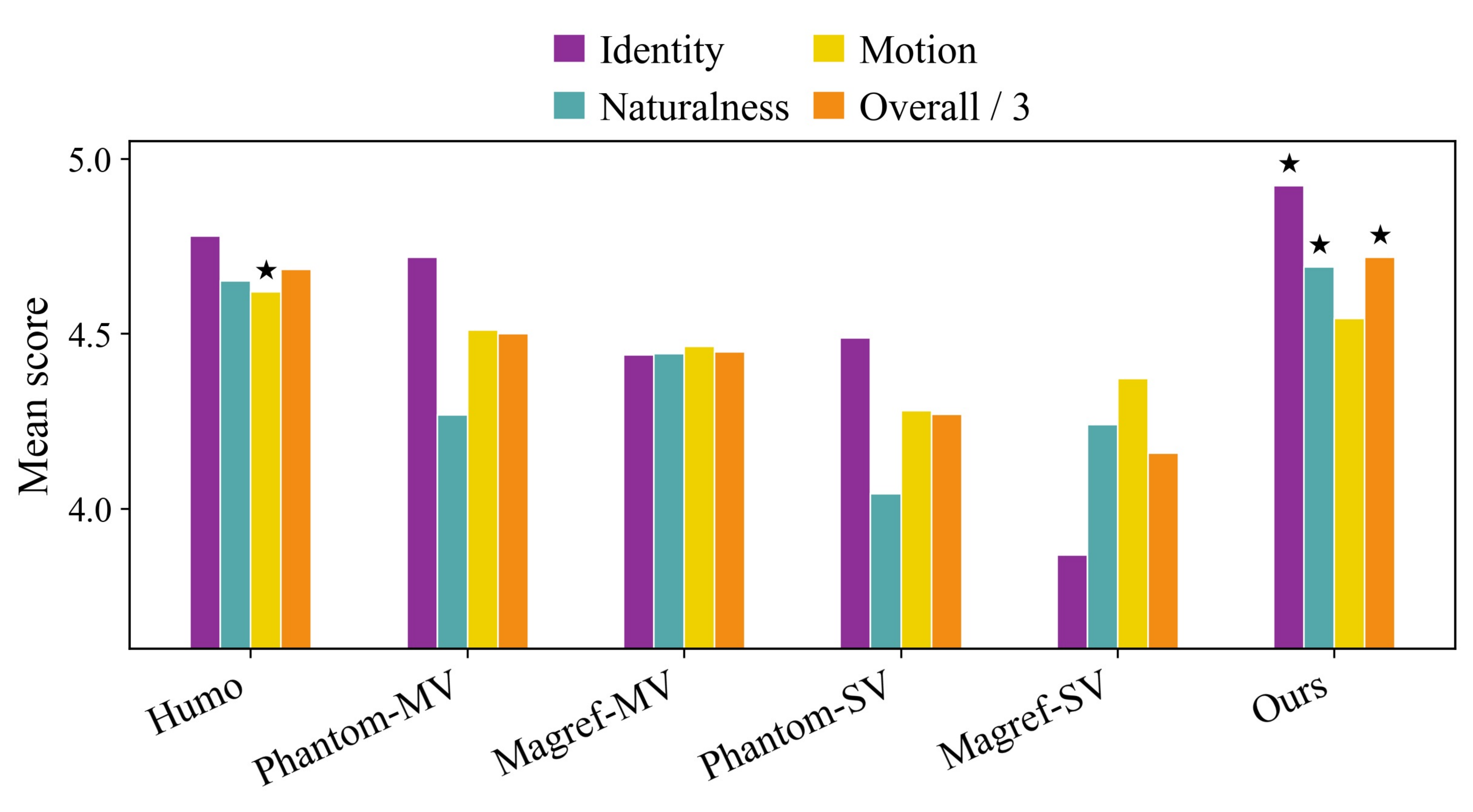}
    \caption{User-study results. }
    \label{fig:user_study}
\end{figure}

We conduct a human study on 25 randomly sampled cases across six methods. 
Participants are shown ten reference images and one anonymized generated video, and rate \emph{Identity Consistency}, \emph{Naturalness}, and \emph{Motion/Viewpoint Naturalness} on a 1--5 Likert scale. 
We report the mean score for each criterion and their average as the overall score (10 evaluator per case). As shown in Fig.~\ref{fig:user_study}, our method achieves the highest scores on \emph{Identity Consistency} and \emph{Naturalness}, while HuMo performs slightly better on \emph{Motion/Viewpoint Naturalness}. 
Overall, our method obtains the best total score, indicating a better balance between identity fidelity and perceptual quality. For \emph{Identity Consistency}, we further conduct statistical tests. Both one-way ANOVA~\cite{fisher1925statistical} and the Kruskal--Wallis test~\cite{kruskal1952use} show significant differences among methods (ANOVA: \(F=81.70\), \(p<10^{-70}\); Kruskal--Wallis: \(H=283.75\), \(p<10^{-50}\)).

\section{Discussion}
We further discuss our work from three aspects.

\textbf{Applications.} Our method is particularly useful for scenarios requiring strong identity consistency under large pose variations, such as recreating known identities as digital characters in films and visual effects.

\textbf{Evaluation of naturalness.} Evaluating facial motion dynamics remains challenging. Simple statistics (e.g., facial angle variance or deviation from reference views) cannot fully capture temporal coherence and realism. Although multimodal models provide automatic evaluation, human evaluation remains the most reliable approach.

\textbf{Balance between consistency and naturalness.} Identity consistency and motion naturalness often exhibit a trade-off: stronger reference constraints improve identity fidelity but may reduce motion diversity. Achieving both remains an open challenge.

\section{Conclusion}
In this work, we study the identity consistency under large facial-angle variations, a setting largely overlooked by prior video generation methods. We propose $\mathrm{Mv}^2\mathrm{ID}$, a human-centric video generation framework based on multi-view conditioning. We introduce a region masking training strategy and a reference-decoupled RoPE to mitigate the view-dependent copy-paste issue under in-paired supervision. We also develop a large-scale data construction pipeline, resulting in a dataset of 22K high-quality videos. Extensive experiments show that our approach achieves a better balance between identity consistency and facial motion naturalness compared to the methods under cross-paired supervision.

\bibliographystyle{splncs04}
\bibliography{main}

@String(CVPR  = {IEEE Conf. Comput. Vis. Pattern Recog.})

@String(ICCV  = {Int. Conf. Comput. Vis.})

@String(NeurIPS = {Adv. Neural Inform. Process. Syst.})

@String(ICML  = {Int. Conf. Mach. Learn.})

@String(CVPR  = {CVPR})

@String(ICCV  = {ICCV})

@String(NeurIPS = {NeurIPS})

@String(ICML  = {ICML})

@inproceedings{arcface2018,
  author    = {Deng, Jiankang and Guo, Jia and Xue, Niannan and Zafeiriou, Stefanos},
  title     = {ArcFace: Additive Angular Margin Loss for Deep Face Recognition},
  booktitle = {Proceedings of the IEEE/CVF Conference on Computer Vision and Pattern Recognition (CVPR)},
  year      = {2019},
  pages     = {4690--4699},
  doi       = {10.1109/CVPR.2019.00482},
  url       = {https://openaccess.thecvf.com/content_CVPR_2019/html/Deng_ArcFace_Additive_Angular_Margin_Loss_for_Deep_Face_Recognition_CVPR_2019_paper.html}
}

@misc{cogvideo2022,
      title={CogVideo: Large-scale Pretraining for Text-to-Video Generation via Transformers}, 
      author={Wenyi Hong and Ming Ding and Wendi Zheng and Xinghan Liu and Jie Tang},
      year={2022},
      eprint={2205.15868},
      archivePrefix ={arXiv},
      primaryClass ={cs.CV},
      url={https://arxiv.org/abs/2205.15868}, 
}

@inproceedings{ddpm2020,
  author    = {Ho, Jonathan and Jain, Ajay and Abbeel, Pieter},
  title     = {Denoising Diffusion Probabilistic Models},
  booktitle = {Advances in Neural Information Processing Systems},
  year      = {2020},
  volume    = {33},
  pages     = {6840--6851},
  publisher = {Curran Associates, Inc.},
  url       = {https://proceedings.neurips.cc/paper_files/paper/2020/file/4c5bcfec8584af0d967f1ab10179ca4b-Paper.pdf}
}

@misc{dreamfusion2022,
  author        = {Poole, Ben and Jain, Ajay and Barron, Jonathan T. and Mildenhall, Ben},
  title         = {DreamFusion: Text-to-3D using 2D Diffusion},
  year          = {2022},
  url           = {https://arxiv.org/abs/2209.14988},
  eprint        = {2209.14988},
  archivePrefix = {arXiv},
  primaryClass = {cs.CV}
}

@inproceedings{edm2022,
  author    = {Tero Karras and
               Miika Aittala and
               Timo Aila and
               Samuli Laine},
  title     = {Elucidating the Design Space of Diffusion-Based Generative Models},
  booktitle = {Advances in Neural Information Processing Systems 35: Annual Conference
               on Neural Information Processing Systems 2022, NeurIPS 2022, New Orleans,
               LA, USA, November 28 - December 9, 2022},
  year      = {2022},
  url       = {http://papers.nips.cc/paper_files/paper/2022/hash/a98846e9d9cc01cfb87eb694d946ce6b-Abstract-Conference.html}
}

@book{fisher1925statistical,
  author    = {Fisher, Ronald Aylmer},
  title     = {Statistical Methods for Research Workers},
  year      = {1925},
  publisher = {Oliver and Boyd},
  address   = {Edinburgh and London}
}

@inproceedings{flow_matching2023,
  author    = {Lipman, Yaron and Chen, Ricky T. Q. and Ben-Hamu, Heli and Nickel, Maximilian and Le, Matthew},
  title     = {Flow Matching for Generative Modeling},
  booktitle = {International Conference on Learning Representations},
  year      = {2023},
  url       = {https://openreview.net/forum?id=PqvMRDCJT9t}
}

@misc{hunyuanvideo15_2025,
      title={HunyuanVideo 1.5 Technical Report}, 
      author={Bing Wu and Chang Zou and Changlin Li and Duojun Huang and Fang Yang and Hao Tan and Jack Peng and Jianbing Wu and Jiangfeng Xiong and Jie Jiang and Linus and Patrol and Peizhen Zhang and Peng Chen and Penghao Zhao and Qi Tian and Songtao Liu and Weijie Kong and Weiyan Wang and Xiao He and Xin Li and Xinchi Deng and Xuefei Zhe and Yang Li and Yanxin Long and Yuanbo Peng and Yue Wu and Yuhong Liu and Zhenyu Wang and Zuozhuo Dai and Bo Peng and Coopers Li and Gu Gong and Guojian Xiao and Jiahe Tian and Jiaxin Lin and Jie Liu and Jihong Zhang and Jiesong Lian and Kaihang Pan and Lei Wang and Lin Niu and Mingtao Chen and Mingyang Chen and Mingzhe Zheng and Miles Yang and Qiangqiang Hu and Qi Yang and Qiuyong Xiao and Runzhou Wu and Ryan Xu and Rui Yuan and Shanshan Sang and Shisheng Huang and Siruis Gong and Shuo Huang and Weiting Guo and Xiang Yuan and Xiaojia Chen and Xiawei Hu and Wenzhi Sun and Xiele Wu and Xianshun Ren and Xiaoyan Yuan and Xiaoyue Mi and Yepeng Zhang and Yifu Sun and Yiting Lu and Yitong Li and You Huang and Yu Tang and Yixuan Li and Yuhang Deng and Yuan Zhou and Zhichao Hu and Zhiguang Liu and Zhihe Yang and Zilin Yang and Zhenzhi Lu and Zixiang Zhou and Zhao Zhong},
      year={2025},
      eprint={2511.18870},
      archivePrefix ={arXiv},
      primaryClass ={cs.CV},
      url={https://arxiv.org/abs/2511.18870}, 
}

@misc{imagen_video2022,
  author        = {Ho, Jonathan and Chan, William and Saharia, Chitwan and Whang, Jay and Gao, Ruiqi and Gritsenko, Alexey and Kingma, Diederik P. and Poole, Ben and Norouzi, Mohammad and Fleet, David J. and Salimans, Tim},
  title         = {Imagen Video: High Definition Video Generation with Diffusion Models},
  year          = {2022},
  url           = {https://arxiv.org/abs/2210.02303},
  eprint        = {2210.02303},
  archivePrefix = {arXiv},
  primaryClass = {cs.CV}
}

@misc{imagereward2023,
      title={ImageReward: Learning and Evaluating Human Preferences for Text-to-Image Generation}, 
      author={Jiazheng Xu and Xiao Liu and Yuchen Wu and Yuxuan Tong and Qinkai Li and Ming Ding and Jie Tang and Yuxiao Dong},
      year={2023},
      eprint={2304.05977},
      archivePrefix ={arXiv},
      primaryClass ={cs.CV},
      url={https://arxiv.org/abs/2304.05977}, 
}

@inproceedings{improved_ddpm2021,
  author    = {Alexander Quinn Nichol and
               Prafulla Dhariwal},
  title     = {Improved Denoising Diffusion Probabilistic Models},
  booktitle = {Proceedings of the 38th International Conference on Machine Learning,
               {ICML} 2021, 18-24 July 2021, Virtual Event},
  series    = {Proceedings of Machine Learning Research},
  volume    = {139},
  pages     = {8162--8171},
  publisher = {{PMLR}},
  year      = {2021},
  url       = {https://proceedings.mlr.press/v139/nichol21a.html}
}

@inproceedings{jiang2024videobooth,
  title={Videobooth: Diffusion-based video generation with image prompts},
  author={Jiang, Yuming and Wu, Tianxing and Yang, Shuai and Si, Chenyang and Lin, Dahua and Qiao, Yu and Loy, Chen Change and Liu, Ziwei},
  booktitle={Proceedings of the IEEE/CVF Conference on Computer Vision and Pattern Recognition},
  pages={6689--6700},
  year={2024}
}

@article{kruskal1952use,
  author  = {Kruskal, William H. and Wallis, W. Allen},
  title   = {Use of Ranks in One-Criterion Variance Analysis},
  journal = {Journal of the American Statistical Association},
  volume  = {47},
  number  = {260},
  pages   = {583--621},
  year    = {1952},
  doi     = {10.1080/01621459.1952.10483441}
}

@inproceedings{ldm2022,
  author    = {Rombach, Robin and Blattmann, Andreas and Lorenz, Dominik and Esser, Patrick and Ommer, Bj{\"o}rn},
  title     = {High-Resolution Image Synthesis With Latent Diffusion Models},
  booktitle = {Proceedings of the IEEE/CVF Conference on Computer Vision and Pattern Recognition (CVPR)},
  year      = {2022},
  pages     = {10684--10695},
  doi       = {10.1109/CVPR52688.2022.01042},
  url       = {https://openaccess.thecvf.com/content/CVPR2022/html/Rombach_High-Resolution_Image_Synthesis_With_Latent_Diffusion_Models_CVPR_2022_paper.html}
}

@inproceedings{Li_2025_ICCV,
  author    = {Li, Hengjia and Qiu, Haonan and Zhang, Shiwei and Wang, Xiang and Wei, Yujie and Li, Zekun and Zhang, Yingya and Wu, Boxi and Cai, Deng},
  title     = {PersonalVideo: High ID-Fidelity Video Customization without Dynamic and Semantic Degradation},
  booktitle = {Proceedings of the IEEE/CVF International Conference on Computer Vision (ICCV)},
  month     = {October},
  year      = {2025},
  pages     = {19406--19416}
}

@inproceedings{Liu_2025_ICCV,
  author    = {Liu, Lijie and Ma, Tianxiang and Li, Bingchuan and Chen, Zhuowei and Liu, Jiawei and Li, Gen and Zhou, Siyu and He, Qian and Wu, Xinglong},
  title     = {Phantom: Subject-Consistent Video Generation via Cross-Modal Alignment},
  booktitle = {Proceedings of the IEEE/CVF International Conference on Computer Vision (ICCV)},
  month     = {October},
  year      = {2025},
  pages     = {14951--14961}
}

@misc{lumiere2024,
      title={Lumiere: A Space-Time Diffusion Model for Video Generation}, 
      author={Omer Bar-Tal and Hila Chefer and Omer Tov and Charles Herrmann and Roni Paiss and Shiran Zada and Ariel Ephrat and Junhwa Hur and Guanghui Liu and Amit Raj and Yuanzhen Li and Michael Rubinstein and Tomer Michaeli and Oliver Wang and Deqing Sun and Tali Dekel and Inbar Mosseri},
      year={2024},
      eprint={2401.12945},
      archivePrefix ={arXiv},
      primaryClass ={cs.CV},
      url={https://arxiv.org/abs/2401.12945}, 
}

@inproceedings{mae2021,
  author    = {He, Kaiming and Chen, Xinlei and Xie, Saining and Li, Yanghao and Doll{\'a}r, Piotr and Girshick, Ross},
  title     = {Masked Autoencoders Are Scalable Vision Learners},
  booktitle = {Proceedings of the IEEE/CVF Conference on Computer Vision and Pattern Recognition (CVPR)},
  year      = {2022},
  pages     = {16000--16009},
  doi       = {10.1109/CVPR52688.2022.01553},
  url       = {https://openaccess.thecvf.com/content/CVPR2022/html/He_Masked_Autoencoders_Are_Scalable_Vision_Learners_CVPR_2022_paper.html}
}

@misc{magic3d2022,
      title={Magic3D: High-Resolution Text-to-3D Content Creation}, 
      author={Chen-Hsuan Lin and Jun Gao and Luming Tang and Towaki Takikawa and Xiaohui Zeng and Xun Huang and Karsten Kreis and Sanja Fidler and Ming-Yu Liu and Tsung-Yi Lin},
      year={2023},
      eprint={2211.10440},
      archivePrefix ={arXiv},
      primaryClass ={cs.CV},
      url={https://arxiv.org/abs/2211.10440}, 
}

@misc{make_a_video2022,
      title={Make-A-Video: Text-to-Video Generation without Text-Video Data}, 
      author={Uriel Singer and Adam Polyak and Thomas Hayes and Xi Yin and Jie An and Songyang Zhang and Qiyuan Hu and Harry Yang and Oron Ashual and Oran Gafni and Devi Parikh and Sonal Gupta and Yaniv Taigman},
      year={2022},
      eprint={2209.14792},
      archivePrefix ={arXiv},
      primaryClass ={cs.CV},
      url={https://arxiv.org/abs/2209.14792}, 
}

@article{talebi2018nima,
  title={NIMA: Neural image assessment},
  author={Talebi, Hossein and Milanfar, Peyman},
  journal={IEEE transactions on image processing},
  volume={27},
  number={8},
  pages={3998--4011},
  year={2018},
  publisher={IEEE}
}

@misc{opensora2024,
      title={Open-Sora: Democratizing Efficient Video Production for All}, 
      author={Zangwei Zheng and Xiangyu Peng and Tianji Yang and Chenhui Shen and Shenggui Li and Hongxin Liu and Yukun Zhou and Tianyi Li and Yang You},
      year={2024},
      eprint={2412.20404},
      archivePrefix ={arXiv},
      primaryClass ={cs.CV},
      url={https://arxiv.org/abs/2412.20404}, 
}

@misc{opensora_plan2024,
      title={Open-Sora Plan: Open-Source Large Video Generation Model}, 
      author={Bin Lin and Yunyang Ge and Xinhua Cheng and Zongjian Li and Bin Zhu and Shaodong Wang and Xianyi He and Yang Ye and Shenghai Yuan and Liuhan Chen and Tanghui Jia and Junwu Zhang and Zhenyu Tang and Yatian Pang and Bin She and Cen Yan and Zhiheng Hu and Xiaoyi Dong and Lin Chen and Zhang Pan and Xing Zhou and Shaoling Dong and Yonghong Tian and Li Yuan},
      year={2024},
      eprint={2412.00131},
      archivePrefix ={arXiv},
      primaryClass ={cs.CV},
      url={https://arxiv.org/abs/2412.00131}, 
}

@misc{phantom_data2025,
      title={Phantom-Data : Towards a General Subject-Consistent Video Generation Dataset}, 
      author={Zhuowei Chen and Bingchuan Li and Tianxiang Ma and Lijie Liu and Mingcong Liu and Yi Zhang and Gen Li and Xinghui Li and Siyu Zhou and Qian He and Xinglong Wu},
      year={2025},
      eprint={2506.18851},
      archivePrefix ={arXiv},
      primaryClass ={cs.CV},
      url={https://arxiv.org/abs/2506.18851}, 
}

@misc{phenaki2022,
      title={Phenaki: Variable Length Video Generation From Open Domain Textual Description}, 
      author={Ruben Villegas and Mohammad Babaeizadeh and Pieter-Jan Kindermans and Hernan Moraldo and Han Zhang and Mohammad Taghi Saffar and Santiago Castro and Julius Kunze and Dumitru Erhan},
      year={2022},
      eprint={2210.02399},
      archivePrefix ={arXiv},
      primaryClass ={cs.CV},
      url={https://arxiv.org/abs/2210.02399}, 
}

@misc{prolificdreamer2023,
      title={ProlificDreamer: High-Fidelity and Diverse Text-to-3D Generation with Variational Score Distillation}, 
      author={Zhengyi Wang and Cheng Lu and Yikai Wang and Fan Bao and Chongxuan Li and Hang Su and Jun Zhu},
      year={2023},
      eprint={2305.16213},
      archivePrefix ={arXiv},
      primaryClass ={cs.LG},
      url={https://arxiv.org/abs/2305.16213}, 
}

@misc{rectified_flow2022,
  author        = {Liu, Xingchao and Gong, Chengyue and Liu, Qiang},
  title         = {Flow Straight and Fast: Learning to Generate and Transfer Data with Rectified Flow},
  year          = {2022},
  url           = {https://arxiv.org/abs/2209.03003},
  eprint        = {2209.03003},
  archivePrefix = {arXiv},
  primaryClass = {cs.LG}
}

@misc{retinaface2019,
      title={RetinaFace: Single-stage Dense Face Localisation in the Wild}, 
      author={Jiankang Deng and Jia Guo and Yuxiang Zhou and Jinke Yu and Irene Kotsia and Stefanos Zafeiriou},
      year={2019},
      eprint={1905.00641},
      archivePrefix ={arXiv},
      primaryClass ={cs.CV},
      url={https://arxiv.org/abs/1905.00641}, 
}

@inproceedings{score_sde2021,
  author    = {Song, Yang and Sohl-Dickstein, Jascha and Kingma, Diederik P and Kumar, Abhishek and Ermon, Stefano and Poole, Ben},
  title     = {Score-Based Generative Modeling through Stochastic Differential Equations},
  booktitle = {International Conference on Learning Representations},
  year      = {2021},
  url       = {https://openreview.net/forum?id=PxTIG12RRHS}
}

@misc{shortcut_mitigation_diffusion2023,
      title={Leveraging Diffusion Disentangled Representations to Mitigate Shortcuts in Underspecified Visual Tasks}, 
      author={Luca Scimeca and Alexander Rubinstein and Armand Mihai Nicolicioiu and Damien Teney and Yoshua Bengio},
      year={2023},
      eprint={2310.02230},
      archivePrefix ={arXiv},
      primaryClass ={cs.CV},
      url={https://arxiv.org/abs/2310.02230}, 
}

@article{t5_2020,
  author  = {Raffel, Colin and Shazeer, Noam and Roberts, Adam and Lee, Katherine and Narang, Sharan and Matena, Michael and Zhou, Yanqi and Li, Wei and Liu, Peter J.},
  title   = {Exploring the Limits of Transfer Learning with a Unified Text-to-Text Transformer},
  journal = {Journal of Machine Learning Research},
  year    = {2020},
  volume  = {21},
  number  = {140},
  pages   = {1--67},
  url     = {http://jmlr.org/papers/v21/20-074.html}
}

@misc{vace2025,
      title={VACE: All-in-One Video Creation and Editing}, 
      author={Zeyinzi Jiang and Zhen Han and Chaojie Mao and Jingfeng Zhang and Yulin Pan and Yu Liu},
      year={2025},
      eprint={2503.07598},
      archivePrefix ={arXiv},
      primaryClass ={cs.CV},
      url={https://arxiv.org/abs/2503.07598}, 
}

@misc{vbench2023,
  author        = {Huang, Ziqi and He, Yinan and Yu, Jiashuo and Zhang, Fan and Si, Chenyang and Jiang, Yuming and Zhang, Yuanhan and Wu, Tianxing and Jin, Qingyang and Chanpaisit, Nattapol and Wang, Yaohui and Chen, Xinyuan and Wang, Limin and Lin, Dahua and Qiao, Yu and Liu, Ziwei},
  title         = {VBench: Comprehensive Benchmark Suite for Video Generative Models},
  year          = {2023},
  url           = {https://arxiv.org/abs/2311.17982},
  eprint        = {2311.17982},
  archivePrefix = {arXiv},
  primaryClass = {cs.CV}
}

@misc{video_diffusion_2022,
  author        = {Ho, Jonathan and Salimans, Tim and Gritsenko, Alexey and Chan, William and Norouzi, Mohammad and Fleet, David J.},
  title         = {Video Diffusion Models},
  year          = {2022},
  url           = {https://arxiv.org/abs/2204.03458},
  eprint        = {2204.03458},
  archivePrefix = {arXiv},
  primaryClass = {cs.CV}
}

@misc{wan2025,
      title={Wan: Open and Advanced Large-Scale Video Generative Models}, 
      author={Team Wan and Ang Wang and Baole Ai and Bin Wen and Chaojie Mao and Chen-Wei Xie and Di Chen and Feiwu Yu and Haiming Zhao and Jianxiao Yang and Jianyuan Zeng and Jiayu Wang and Jingfeng Zhang and Jingren Zhou and Jinkai Wang and Jixuan Chen and Kai Zhu and Kang Zhao and Keyu Yan and Lianghua Huang and Mengyang Feng and Ningyi Zhang and Pandeng Li and Pingyu Wu and Ruihang Chu and Ruili Feng and Shiwei Zhang and Siyang Sun and Tao Fang and Tianxing Wang and Tianyi Gui and Tingyu Weng and Tong Shen and Wei Lin and Wei Wang and Wei Wang and Wenmeng Zhou and Wente Wang and Wenting Shen and Wenyuan Yu and Xianzhong Shi and Xiaoming Huang and Xin Xu and Yan Kou and Yangyu Lv and Yifei Li and Yijing Liu and Yiming Wang and Yingya Zhang and Yitong Huang and Yong Li and You Wu and Yu Liu and Yulin Pan and Yun Zheng and Yuntao Hong and Yupeng Shi and Yutong Feng and Zeyinzi Jiang and Zhen Han and Zhi-Fan Wu and Ziyu Liu},
      year={2025},
      eprint={2503.20314},
      archivePrefix ={arXiv},
      primaryClass ={cs.CV},
      url={https://arxiv.org/abs/2503.20314}, 
}

@inproceedings{Wei_2024_CVPR,
  author    = {Wei, Yujie and Zhang, Shiwei and Qing, Zhiwu and Yuan, Hangjie and Liu, Zhiheng and Liu, Yu and Zhang, Yingya and Zhou, Jingren and Shan, Hongming},
  title     = {DreamVideo: Composing Your Dream Videos with Customized Subject and Motion},
  booktitle = {Proceedings of the IEEE/CVF Conference on Computer Vision and Pattern Recognition (CVPR)},
  month     = {June},
  year      = {2024},
  pages     = {6537--6549}
}

@inproceedings{Wu_2023_ICCV,
  author    = {Wu, Jay Zhangjie and Ge, Yixiao and Wang, Xintao and Lei, Stan Weixian and Gu, Yuchao and Shi, Yufei and Hsu, Wynne and Shan, Ying and Qie, Xiaohu and Shou, Mike Zheng},
  title     = {Tune-A-Video: One-Shot Tuning of Image Diffusion Models for Text-to-Video Generation},
  booktitle = {Proceedings of the IEEE/CVF International Conference on Computer Vision (ICCV)},
  month     = {October},
  year      = {2023},
  pages     = {7623--7633}
}

@inproceedings{yuan2025opensvnexus,
  title     = {OpenS2V-Nexus: A Detailed Benchmark and Million-Scale Dataset for Subject-to-Video Generation},
  author    = {Yuan, Shenghai and He, Xianyi and Deng, Yufan and Ye, Yang and Huang, Jinfa and Lin, Bin and Ma, Chongyang and Luo, Jiebo and Yuan, Li},
  booktitle = {The Thirty-ninth Annual Conference on Neural Information Processing Systems Datasets and Benchmarks Track},
  year      = {2025},
  url       = {https://openreview.net/forum?id=XKhLsRPMsw}
}

@misc{zero1232023,
      title={Zero-1-to-3: Zero-shot One Image to 3D Object}, 
      author={Ruoshi Liu and Rundi Wu and Basile Van Hoorick and Pavel Tokmakov and Sergey Zakharov and Carl Vondrick},
      year={2023},
      eprint={2303.11328},
      archivePrefix ={arXiv},
      primaryClass ={cs.CV},
      url={https://arxiv.org/abs/2303.11328}, 
}

@article{Zhang2025Kaleido,
  title         = {Kaleido: Open-Sourced Multi-Subject Reference Video Generation Model},
  author        = {Zhang, Zhenxing and Teng, Jiayan and Yang, Zhuoyi and Cao, Tiankun and Wang, Cheng and Gu, Xiaotao and Tang, Jie and Guo, Dan and Wang, Meng},
  year          = {2025},
  eprint        = {2510.18573},
  archivePrefix = {arXiv},
  primaryClass  = {cs.CV}
}

@article{Deng2025MAGREF,
  title={MAGREF: Masked Guidance for Any-Reference Video Generation with Subject Disentanglement},
  author={Deng, Yufan and Yin, Yuanyang and Guo, Xun and Wang, Yizhi and Fang, Jacob Zhiyuan and Yuan, Shenghai and Yang, Yiding and Wang, Angtian and Liu, Bo and Huang, Haibin and others},
  journal={arXiv preprint arXiv:2505.23742},
  year={2025}
}

@misc{chen2025humo,
      title={HuMo: Human-Centric Video Generation via Collaborative Multi-Modal Conditioning}, 
      author={Liyang Chen and Tianxiang Ma and Jiawei Liu and Bingchuan Li and Zhuowei Chen and Lijie Liu and Xu He and Gen Li and Qian He and Zhiyong Wu},
      year={2025},
      eprint={2509.08519},
      archivePrefix={arXiv},
      primaryClass={cs.CV},
      url={https://arxiv.org/abs/2509.08519}, 
}

@misc{qwen2025qwen25technicalreport,
  title         = {Qwen2.5 Technical Report},
  author        = {Qwen and Yang, An and Yang, Baosong and Zhang, Beichen and Hui, Binyuan and Zheng, Bo and Yu, Bowen and Li, Chengyuan and Liu, Dayiheng and Huang, Fei and Wei, Haoran and Lin, Huan and Yang, Jian and Tu, Jianhong and Zhang, Jianwei and Yang, Jianxin and Yang, Jiaxi and Zhou, Jingren and Lin, Junyang and Dang, Kai and Lu, Keming and Bao, Keqin and Yang, Kexin and Yu, Le and Li, Mei and Xue, Mingfeng and Zhang, Pei and Zhu, Qin and Men, Rui and Lin, Runji and Li, Tianhao and Tang, Tianyi and Xia, Tingyu and Ren, Xingzhang and Ren, Xuancheng and Fan, Yang and Su, Yang and Zhang, Yichang and Wan, Yu and Liu, Yuqiong and Cui, Zeyu and Zhang, Zhenru and Qiu, Zihan},
  year          = {2025},
  eprint        = {2412.15115},
  archivePrefix = {arXiv},
  primaryClass  = {cs.CL},
  url           = {https://arxiv.org/abs/2412.15115}
}

@article{Liu2024DiffShortcut,
  title         = {Rethinking and Red-Teaming Protective Perturbation in Personalized Diffusion Models},
  author        = {Liu, Yixin and Chen, Ruoxi and Chen, Xun and Sun, Lichao},
  year          = {2024},
  eprint        = {2406.18944},
  archivePrefix = {arXiv},
  primaryClass  = {cs.CV},
  doi           = {10.48550/arXiv.2406.18944},
  url           = {https://arxiv.org/abs/2406.18944}
}

@misc{openai2024gpt4o,
  author       = {OpenAI},
  title        = {GPT-4o System Card},
  year         = {2024},
  howpublished = {\url{https://openai.com/index/gpt-4o-system-card/}},
  note         = {Accessed: 2026-03-04}
}

@misc{bytedance2026seed2,
  author       = {ByteDance Seed Team},
  title        = {Seed2.0 (Model Page and Model Card)},
  year         = {2026},
  howpublished = {\url{https://seed.bytedance.com/zh/seed2}},
  note         = {Accessed: 2026-03-04}
}

@inproceedings{qi2024layered,
  title={Layered rendering diffusion model for controllable zero-shot image synthesis},
  author={Qi, Zipeng and Huang, Guoxi and Liu, Chenyang and Ye, Fei},
  booktitle={European Conference on Computer Vision},
  pages={426--443},
  year={2024},
  organization={Springer}
}

@article{qi2023difftalker,
  title={Difftalker: Co-driven audio-image diffusion for talking faces via intermediate landmarks},
  author={Qi, Zipeng and Zhang, Xulong and Cheng, Ning and Xiao, Jing and Wang, Jianzong},
  journal={arXiv preprint arXiv:2309.07509},
  year={2023}
}

@misc{song2026mvs2vmultiviewsubjectconsistentvideo,
    title={MV-S2V: Multi-View Subject-Consistent Video Generation}, 
    author={Ziyang Song and Xinyu Gong and Bangya Liu and Zelin Zhao},
    year={2026},
    eprint={2601.17756},
    archivePrefix={arXiv},
    primaryClass={cs.CV},
    url={https://arxiv.org/abs/2601.17756}, 
}

@InProceedings{Huang_2020_CVPR,
  title     = {CurricularFace: Adaptive Curriculum Learning Loss for Deep Face Recognition},
  author    = {Huang, Yuge and Wang, Yuhan and Tai, Ying and Liu, Xiaoming and Shen, Pengcheng and Li, Shaoxin and Li, Jilin and Huang, Feiyue},
  booktitle = {Proceedings of the IEEE/CVF Conference on Computer Vision and Pattern Recognition (CVPR)},
  year      = {2020}
}

@inproceedings{xu2025hunyuanportrait,
  title={Hunyuanportrait: Implicit condition control for enhanced portrait animation},
  author={Xu, Zunnan and Yu, Zhentao and Zhou, Zixiang and Zhou, Jun and Jin, Xiaoyu and Hong, Fa-Ting and Ji, Xiaozhong and Zhu, Junwei and Cai, Chengfei and Tang, Shiyu and others},
  booktitle={Proceedings of the Computer Vision and Pattern Recognition Conference},
  pages={15909--15919},
  year={2025}
}

@inproceedings{hong2025audio,
  title={Audio-visual controlled video diffusion with masked selective state spaces modeling for natural talking head generation},
  author={Hong, Fa-Ting and Xu, Zunnan and Zhou, Zixiang and Zhou, Jun and Li, Xiu and Lin, Qin and Lu, Qinglin and Xu, Dan},
  booktitle={Proceedings of the IEEE/CVF International Conference on Computer Vision (ICCV)},
  year={2025}
}

@inproceedings{zhao2025gshoi,
  title={GSHOI Denoiser: Denoising Gaussian Hand-Object Interaction for Photorealistic Rendering},
  author={Zhao, Lizhi and Lu, Xuequan and Hu, Bin and Ke, Wei and Wang, Lili},
  booktitle={2025 IEEE International Symposium on Mixed and Augmented Reality (ISMAR)},
  pages={614--623},
  year={2025},
  organization={IEEE}
}

@article{huang2026consistentid,
  title={Consistentid: Portrait generation with multimodal fine-grained identity preserving},
  author={Huang, Jiehui and Dong, Xiao and Song, Wenhui and Chong, Zheng and Tang, Zhenchao and Zhou, Jun and Cheng, Yuhao and Chen, Long and Li, Hanhui and Yan, Yiqiang and others},
  journal={IEEE Transactions on Pattern Analysis and Machine Intelligence},
  year={2026},
  publisher={IEEE}
}

\clearpage

\appendix

\vspace{1em}
\begin{center}
    {\Large \bfseries Supplementary Materials}
    
    \vspace{0.5em}
    \rule{0.6\linewidth}{0.4pt}
\end{center}
\vspace{1em}

\section{Additional Analysis of Main Experiments}
\label{sec:expsetting}

\textbf{Computational environment.} The experiments are conducted on a computing cluster with GPUs of NVIDIA\textsuperscript{\textregistered} Tesla\textsuperscript{\texttrademark} A800.

\noindent\textbf{Supervision difference.} In-paired supervision samples reference images and target frames from temporally nearby segments of the same source video, whereas cross-paired supervision pairs reference images with target frames from different videos of the same identity. In practice, cross-paired data are typically constructed by editing in-paired references or collecting images of the same identity from different sources or shots. However, such cross-paired data with large facial-angle variations are difficult to obtain. Therefore, our goal is to balance identity consistency and motion naturalness under low-cost in-paired supervision, while achieving performance comparable to methods trained with cross-paired data.

\subsection{Dataset Construction}
\label{sec:dataset_construction}
\begin{figure}
    \centering
    \includegraphics[width=1\linewidth]{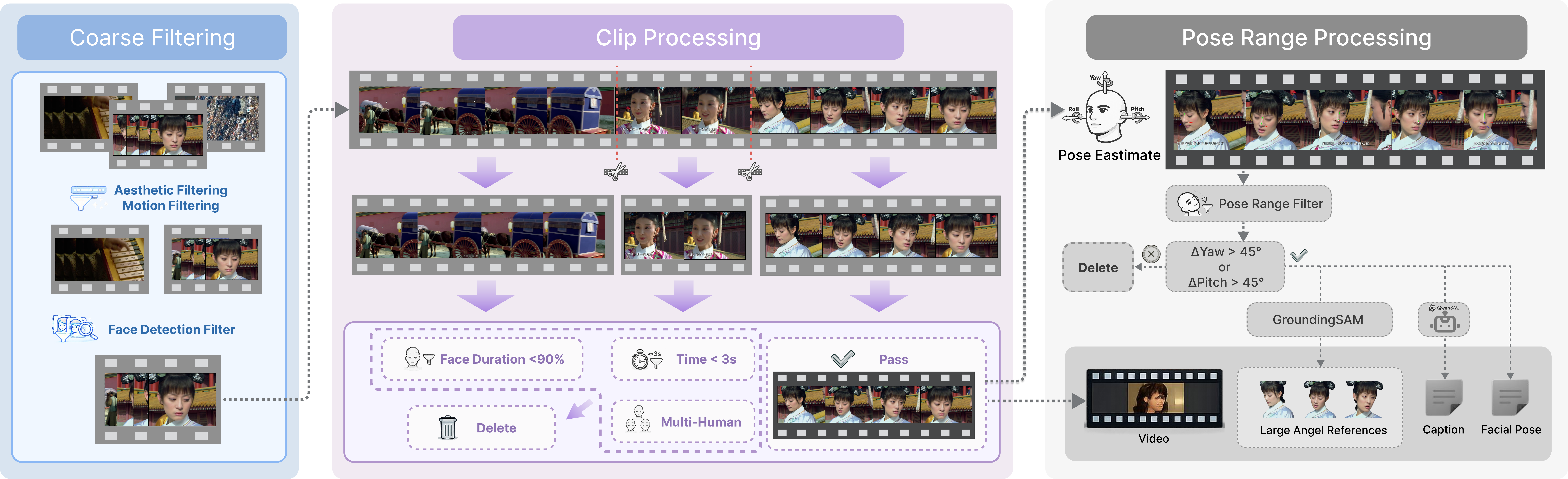}
    \caption{Overview of our dataset construction pipeline. We progressively filter raw videos through coarse filtering, clip-level processing, and pose-level processing to obtain human-centric video clips with large facial-angle variation.}
    \label{fig:dataset_pipeline}
\end{figure}

We provide additional details of the dataset construction pipeline. As illustrated in Fig.~\ref{fig:dataset_pipeline}, the pipeline consists of three stages: coarse filtering, clip-level processing, and pose-level processing.

\textbf{Coarse filtering.}
We first perform initial screening based on visual quality and aesthetic fidelity to remove low-quality or irrelevant videos. To ensure suitability for human-centric facial analysis, we further apply an early face-detection step, discarding videos without visible faces as well as samples with poor facial quality.

\textbf{Clip-level processing.}
For the remaining videos, long videos are segmented into shorter clips. We remove clips shorter than 3 seconds and discard samples containing multiple persons. To ensure reliable facial supervision, we further filter out clips where the target face appears only briefly, retaining samples with sufficiently persistent facial presence.

\textbf{Pose-level processing.}
For each retained clip, we estimate head pose and extract pose attributes, including yaw and pitch. We explicitly retain samples with large pose variation, keeping clips whose pose angle differences exceed $45^\circ$ in either yaw or pitch, while filtering out small-angle cases. To enrich the annotations, we additionally apply Grounding DINO and SAM for segmentation to remove background. The final dataset therefore provides both pose metadata and text descriptions for each sample.

Finally, we randomly sample three frames from each video with pairwise pose differences greater than $45^\circ$, and crop their facial and upper-body regions as reference images.

As summarized in Tab.~\ref{tab:dataset_compare}, existing open-source reference-to-video datasets mainly focus on subject consistency under single-view settings or general human-centric generation, with limited emphasis on large facial-angle variations. In contrast, our dataset is specifically designed for human-centric video generation under large facial-angle variations, with explicit pose filtering and associated pose metadata.

\begin{table}[t]
    \centering
    \footnotesize
    \setlength{\tabcolsep}{4pt}
    \caption{Comparison with representative open-source subject-to-video related datasets. Our dataset explicitly focuses on human-centric samples with large facial-angle variations.}
    \label{tab:dataset_compare}
    \begin{tabular}{l|ccc}
        \toprule
        \textbf{Dataset}
        & \textbf{Human-centric}
        & \textbf{Large-Angle}
        & \textbf{Pose} \\
        \midrule
        OpenS2V-5M~\cite{yuan2025opensvnexus}   & $\times$     & $\times$     & $\times$     \\
        Phantom-Data~\cite{phantom_data2025}    & $\times$     & $\times$     & $\times$     \\
        HuMoSet~\cite{chen2025humo}             & \checkmark   & $\times$     & $\times$     \\
        Ours                                    & \checkmark   & \checkmark   & \checkmark   \\
        \bottomrule
    \end{tabular}
\end{table}

\subsection{Region Masking vs. View Masking}

First, we provide more details on the working principles of view masking and region masking (Section 4.3 in the main text).

\textbf{View Masking.}
The key idea of view masking is to prevent the model from directly attending to the reference view that has the most similar facial pose to the target frame, which could otherwise lead to view-dependent copy-paste behavior. 
Since latent tokens do not explicitly encode pose information, the pose matching is performed in the image space before VAE encoding. 
Specifically, in our video VAE, every four consecutive frames are temporally compressed into one video latent token. For each video latent we use the first frame among its corresponding four frames as the pose anchor. We estimate the facial pose (yaw and pitch) of this anchor frame and all reference images using the same pose estimator. The reference image whose pose is closest to that of the anchor frame is selected as the matched reference view. Let $r_j$ denote this reference image and $R_j = \mathrm{VAE}(r_j)$ its latent representation. During training, we mask the attention between the $i$-th video latent and the matched reference latent by setting
\[
A_{i,j} = 0,
\]
which blocks the most pose-aligned shortcut and encourages the model to aggregate identity information from multiple reference views rather than relying on a single view.

\textbf{Region Masking.}
While view masking blocks the most pose-aligned reference view, the model may still access similar information through indirect attention paths. 
To further reduce the shortcut, we introduce region masking (RM), which randomly masks spatial regions of all reference images before VAE encoding. 
Specifically, a binary mask is applied to each reference image in the pixel space with a fixed masking ratio, and the masked images are then encoded into reference latents. By removing partial appearance cues from every reference view, RM prevents the model from relying on a specific region or view, encouraging it to aggregate identity information across multiple references and improving robustness to view-dependent copy-paste.

\begin{figure}
    \centering
    \includegraphics[width=0.80\linewidth]{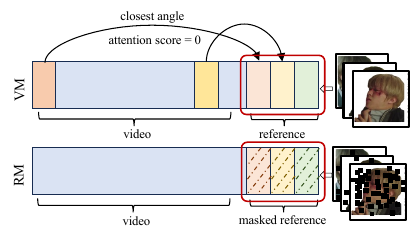}
    \caption{The high-level understanding of VM and RM.}
    \label{fig:vm-rm}
\end{figure}

The high-level understanding of the difference between VM and RM is illustrated in Fig.~\ref{fig:vm-rm}.

\begin{figure}[t]
    \centering
    \includegraphics[width=1\linewidth]{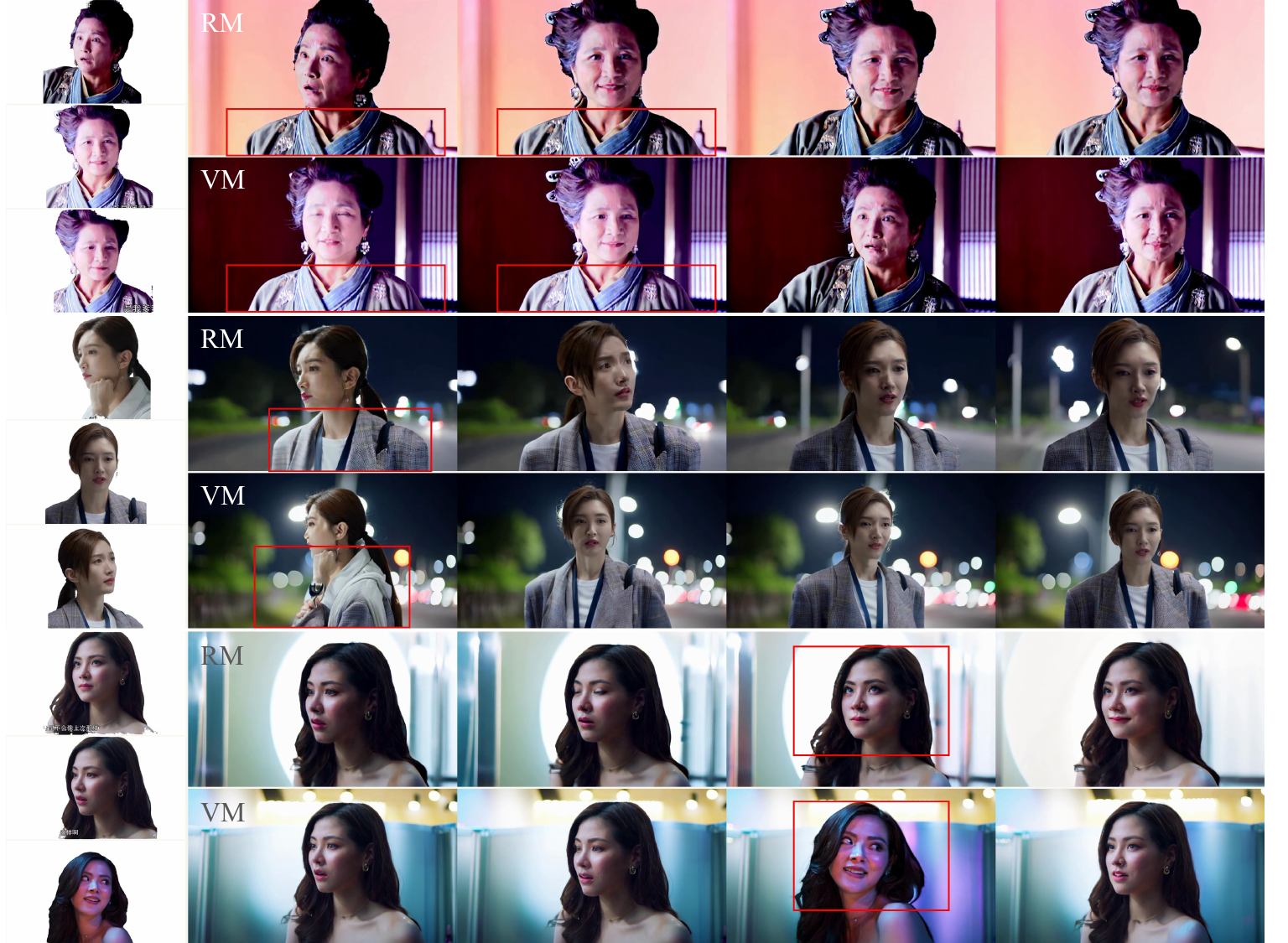}
    \caption{The visualization results of VM and RM. }
    \label{fig:visulization-vm-rm}
\end{figure}

However, VM has two main limitations, as discussed in the main text. These limitations lead to weaker identity consistency and sensitivity to lighting and clothing variations in the reference images, as illustrated in Fig.~\ref{fig:visulization-vm-rm}. This indicates that VM alone has limited ability to mitigate the \textit{copy-paste} issue. In contrast, the visualization results demonstrate that RM is more robust to such variations and achieves stronger identity consistency.

\subsection{Effect of Different Masking Ratios}
\begin{figure}[t]
    \centering
    \includegraphics[width=1\linewidth]{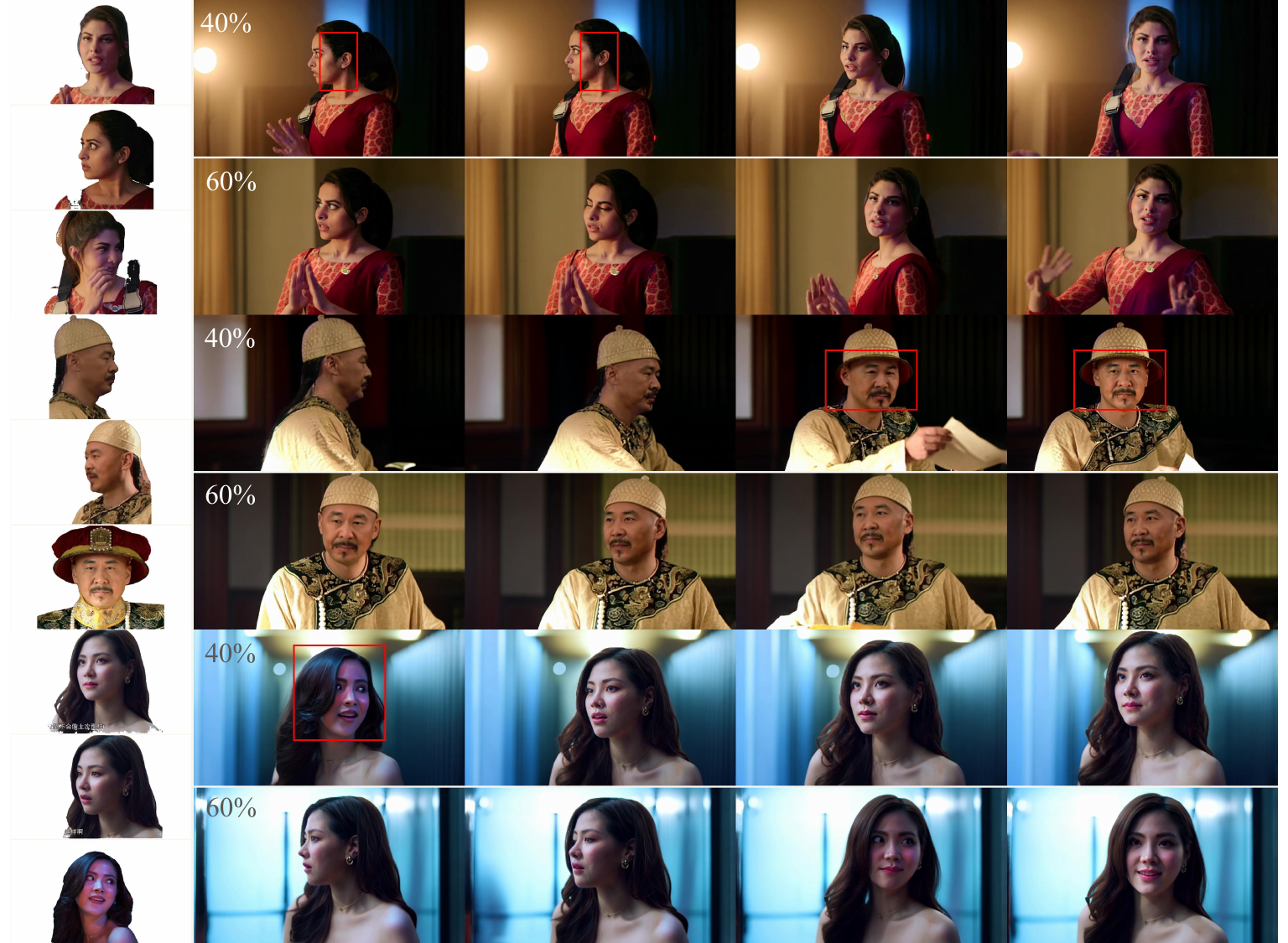}
    \caption{Qualitative comparison of different masking ratios in Region Masking. A higher masking ratio helps mitigate the copy-paste issue and produces more natural results.}
    \label{fig:RMRatio}
\end{figure}

To further analyze the effect of Region Masking, we conduct experiments with two masking ratios: 40\% and 60\%. 
Due to time constraints, we only evaluate these two representative masking ratios.
As shown in Fig.~\ref{fig:RMRatio}, a masking ratio of 40\% still suffers from the copy-paste issue, whereas 60\% alleviates this problem more effectively and produces more natural results. 
We further compare the NaturalScore~\cite{yuan2025opensvnexus} following the OpenS2V-Nexus evaluation protocol. 
As shown in Table~\ref{tab:rm_ratio}, the 60\% setting achieves a higher score (4.30) than the 40\% setting (4.13), indicating that a 60\% masking ratio leads to better visual naturalness in our setting.

\begin{table}[t]
    \centering
    \small
    \setlength{\tabcolsep}{10pt}
    \caption{NaturalScore under different masking ratios. Higher is better.}
    \label{tab:rm_ratio}
    \begin{tabular}{l|c}
        \toprule
        \textbf{Masking Ratio} & \textbf{NaturalScore} \\
        \midrule
        40\% & 4.13 \\
        60\% & \textbf{4.30} \\
        \bottomrule
    \end{tabular}
\end{table}

\subsection{Why Do We Need H/W Offset in RD-RoPE?}
In the proposed reference-decoupled RoPE (RD-RoPE), all reference images share the same frame index since no temporal ordering exists among them. A temporal offset is applied to the reference tokens (see the main text for details). In addition, spatial H/W offsets are introduced to decouple the coordinate systems of video and reference tokens. To better understand the effect of these offsets, we provide a brief mathematical analysis together with visualization comparisons between models trained with and without H/W offsets at an early training stage (e.g., after 450 steps).

Let $q_i^{(m)}, k_j^{(m)} \in \mathbb{R}^2$ denote the $m$-th two-dimensional subspace of the query and key vectors. In RoPE, each token embedding is rotated according to its position $p$:

\begin{equation}
q_i^{(m)'} = R(\omega_m p_i) q_i^{(m)}, \qquad
k_j^{(m)'} = R(\omega_m p_j) k_j^{(m)},
\end{equation}

where

\begin{equation}
R(\theta) =
\begin{pmatrix}
\cos\theta & -\sin\theta \\
\sin\theta & \cos\theta
\end{pmatrix}
\end{equation}

is a 2D rotation matrix and $\omega_m$ denotes the frequency associated with the $m$-th subspace.

The attention score between token $i$ and token $j$ is

\begin{equation}
S_{ij}
=
\sum_m
(q_i^{(m)'})^{\top} k_j^{(m)'}.
\end{equation}

Substituting the RoPE transformation yields

\begin{equation}
S_{ij}
=
\sum_m
(q_i^{(m)})^{\top}
R(\omega_m p_i)^{\top}
R(\omega_m p_j)
k_j^{(m)}.
\end{equation}

Using the identity $R(a)^{\top}R(b)=R(b-a)$, we obtain

\begin{equation}
\label{eq:rope_relative}
S_{ij}
=
\sum_m
(q_i^{(m)})^{\top}
R\bigl(\omega_m (p_j - p_i)\bigr)
k_j^{(m)}.
\end{equation}

Equation~(\ref{eq:rope_relative}) shows that RoPE attention depends only on the relative position $\Delta p = p_j - p_i$.

\paragraph{Temporal Offset Only.}

Assume the token position is represented by $p=(t,x,y)$.
For a reference token $j$ and a video token $i$, the relative position is

\begin{equation}
\Delta p = (\Delta t, \Delta x, \Delta y).
\end{equation}

If only the temporal index is offset,

\begin{equation}
t_r = t^{\text{last}}_{\text{video}} + o_t ,
\end{equation}

the relative temporal position becomes

\begin{equation}
\Delta t = t_r - t_i = (t^{\text{last}}_{\text{video}} - t_i) + o_t .
\end{equation}

In practice, RoPE is typically applied independently along different
positional axes. Therefore each RoPE subspace corresponds to a specific
dimension (temporal, height, or width), and the rotation angle can be written as

\begin{equation}
\theta_m =
\begin{cases}
\omega_m^t \Delta t, & m \in \mathcal{M}_t,\\
\omega_m^x \Delta x, & m \in \mathcal{M}_x,\\
\omega_m^y \Delta y, & m \in \mathcal{M}_y,
\end{cases}
\end{equation}

where $\mathcal{M}_t$, $\mathcal{M}_x$, and $\mathcal{M}_y$ denote the
temporal, height, and width RoPE subspaces, respectively.

Since RoPE employs exponentially spaced frequencies

\begin{equation}
\omega_m = 10000^{-2m/d},
\end{equation}

introducing the temporal offset results in an additional phase shift. Consequently, only the temporal RoPE subspaces are shifted while the
spatial RoPE components remain unchanged.
This breaks the positional distribution learned by the pretrained model,
since the joint positional structure changes from

\begin{equation}
(\Delta t, \Delta x, \Delta y)
\end{equation}

to

\begin{equation}
(\Delta t + o_t, \Delta x, \Delta y).
\end{equation}

Such a positional distribution mismatch makes it more difficult for the
attention mechanism to align reference tokens with video tokens during
the early stages of training. Moreover, the temporal offset introduces frequency-dependent phase
shifts in RoPE. Since RoPE uses exponentially spaced frequencies
$\omega_m = 10000^{-2m/d}$, lower-index subspaces (small $m$) correspond
to higher frequencies and therefore experience larger phase shifts
$\Delta\theta_m = \omega_m o_t$. As a result, the cosine and sine terms
in these subspaces oscillate more rapidly, leading to inconsistent phase
alignment across RoPE dimensions. When aggregating attention
contributions over all subspaces, these inconsistencies may partially
cancel each other, weakening the alignment between reference tokens and
video tokens.

\paragraph{Offset in Spatial Dimensions.}

When spatial offsets are also introduced,

\begin{equation}
x_r \rightarrow x_r + o_x, \qquad
y_r \rightarrow y_r + o_y,
\end{equation}

the relative position becomes

\begin{equation}
\Delta p =
(\Delta t + o_t,\,
\Delta x + o_x,\,
\Delta y + o_y).
\end{equation}

Accordingly, the phase shifts are applied consistently across temporal
and spatial RoPE subspaces:

\begin{equation}
\theta_m =
\begin{cases}
\omega_m^t (\Delta t + o_t), & m \in \mathcal{M}_t,\\
\omega_m^x (\Delta x + o_x), & m \in \mathcal{M}_x,\\
\omega_m^y (\Delta y + o_y), & m \in \mathcal{M}_y.
\end{cases}
\end{equation}

This effectively defines a new but coherent coordinate system.
Since RoPE attention depends only on relative positions,
the positional structure remains balanced across dimensions,
leading to more stable attention alignment between reference tokens
and video tokens.

\paragraph{Discussion.}

Applying offsets jointly to temporal and spatial dimensions preserves
a balanced positional structure in RoPE and effectively decouples
reference tokens from video tokens. Empirically, we observe that this
significantly improves reference conditioning and accelerates model
convergence, as the reference tokens remain distinguishable in
positional space while maintaining stable attention alignment.

\begin{figure}[t]
    \centering
    \includegraphics[width=1\linewidth]{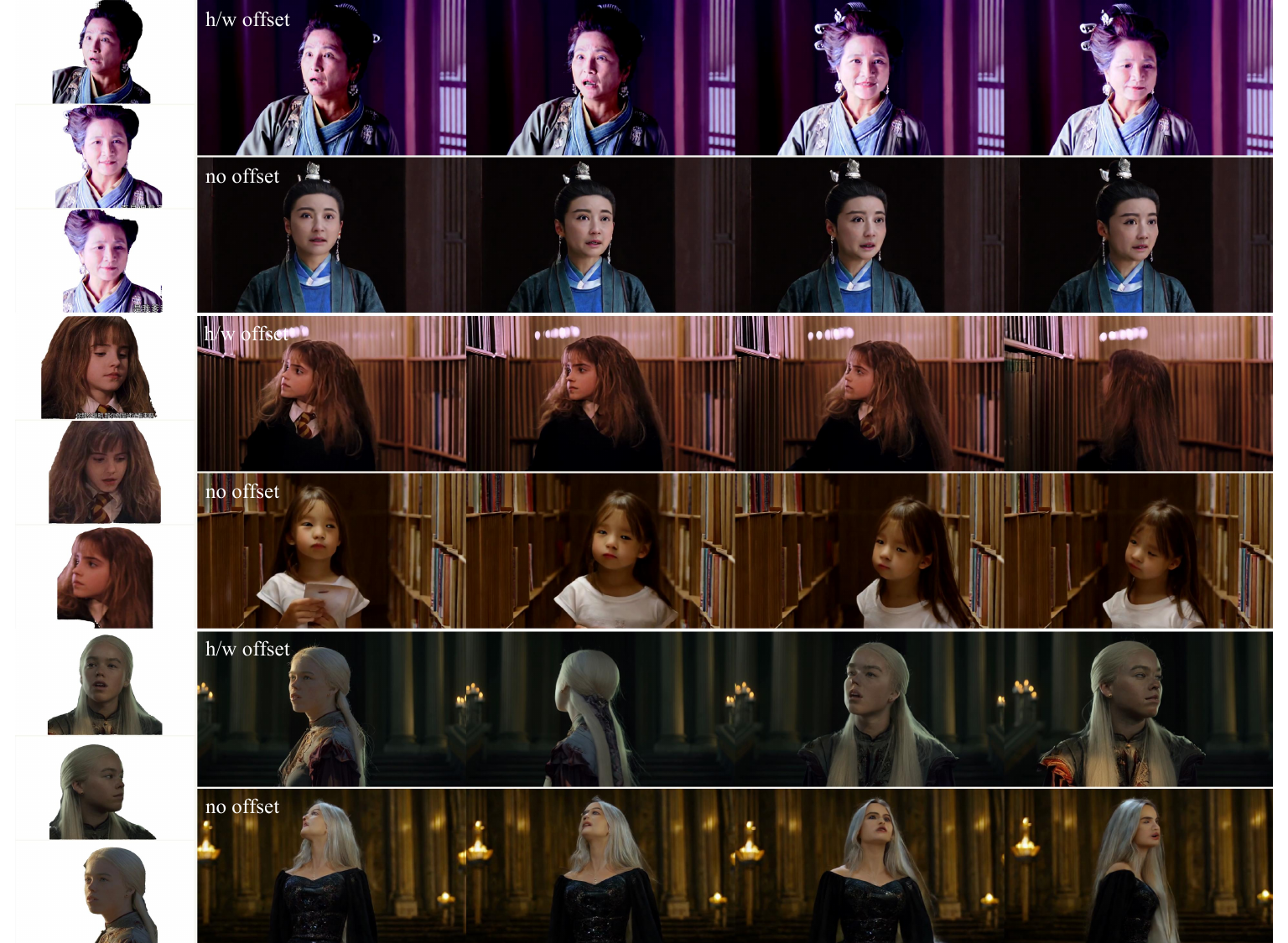}
    \caption{The h/w offset can accelerate the convergence of training. The model with the h/w offset learns to utilize reference information within only 450 training steps.}
    \label{fig:offset}
\end{figure}

The results in Fig.~\ref{fig:offset} further illustrate that, at an early stage of training, the model with H/W offsets learns to utilize reference information more effectively. In contrast, the model without offsets struggles to leverage the reference images, although it can still generate large facial-angle variations and maintain semantic consistency with the text prompt.

\section{The Impact of Number of Reference Images}
\begin{table}[t]
    \centering
    \small
    \setlength{\tabcolsep}{5pt}
    \caption{
    Ablation study of the number of reference images.
    }
    \label{tab:ablation}
    \begin{threeparttable}

    \begin{adjustbox}{max width=\linewidth}
    \begin{tabular}{lccccc}
        \toprule
        \multirow{2}{*}{\textbf{Setting}} 
        & \multicolumn{2}{c}{\textbf{Quality}} 
        & \textbf{Align.} 
        & \multicolumn{2}{c}{\textbf{Identity}} \\
        \cmidrule(lr){2-3}
        \cmidrule(lr){4-4}
        \cmidrule(lr){5-6}
        & \textbf{AES}$\uparrow$
        & \textbf{IQA}$\uparrow$
        & \textbf{TVA}$\uparrow$
        & \textbf{MvRC-Arc}$\uparrow$
        & \textbf{MvRC-Cur}$\uparrow$ \\
        \midrule
        1 Ref & 0.561 & 0.658 & 0.084 & 0.444 & 0.403 \\
        2 Ref & 0.572 & 0.659 & 0.085 & 0.502 & 0.463 \\
        3 Ref & 0.569 & 0.654 & 0.087 & 0.504 & 0.464 \\
        \bottomrule
    \end{tabular}
    \end{adjustbox}

    \begin{tablenotes}[flushleft]
        \footnotesize
        \item \textit{B}: Base method. \textit{R}: RD-RoPE. \textit{M}: Region-Masking Training.
    \end{tablenotes}
    \end{threeparttable}
\end{table}




We further analyze the effect of varying the number of reference views. The experimental results reveal two key observations.

First, a significant improvement in identity consistency is observed when increasing the number of reference views from one to two, which supports our core insight that richer reference information leads to stronger identity consistency(\textit{more information input, stronger consistency get}).

Second, although introducing a third reference view provides a slight improvement, the marginal gain is limited. Therefore, we explore up to three reference views in this work, which also corresponds to the configuration adopted in our main experiments.

Overall, these results suggest that strong performance in large facial-angle variation scenarios can be achieved with only two to three reference views, indicating that extensive multi-view coverage is not strictly required.

\section{More Results}
Here, we provide more visualization and facial trajectory results.
\subsection{More Visualization Results}
\begin{figure}[t]
    \centering
    \includegraphics[width=1\linewidth]{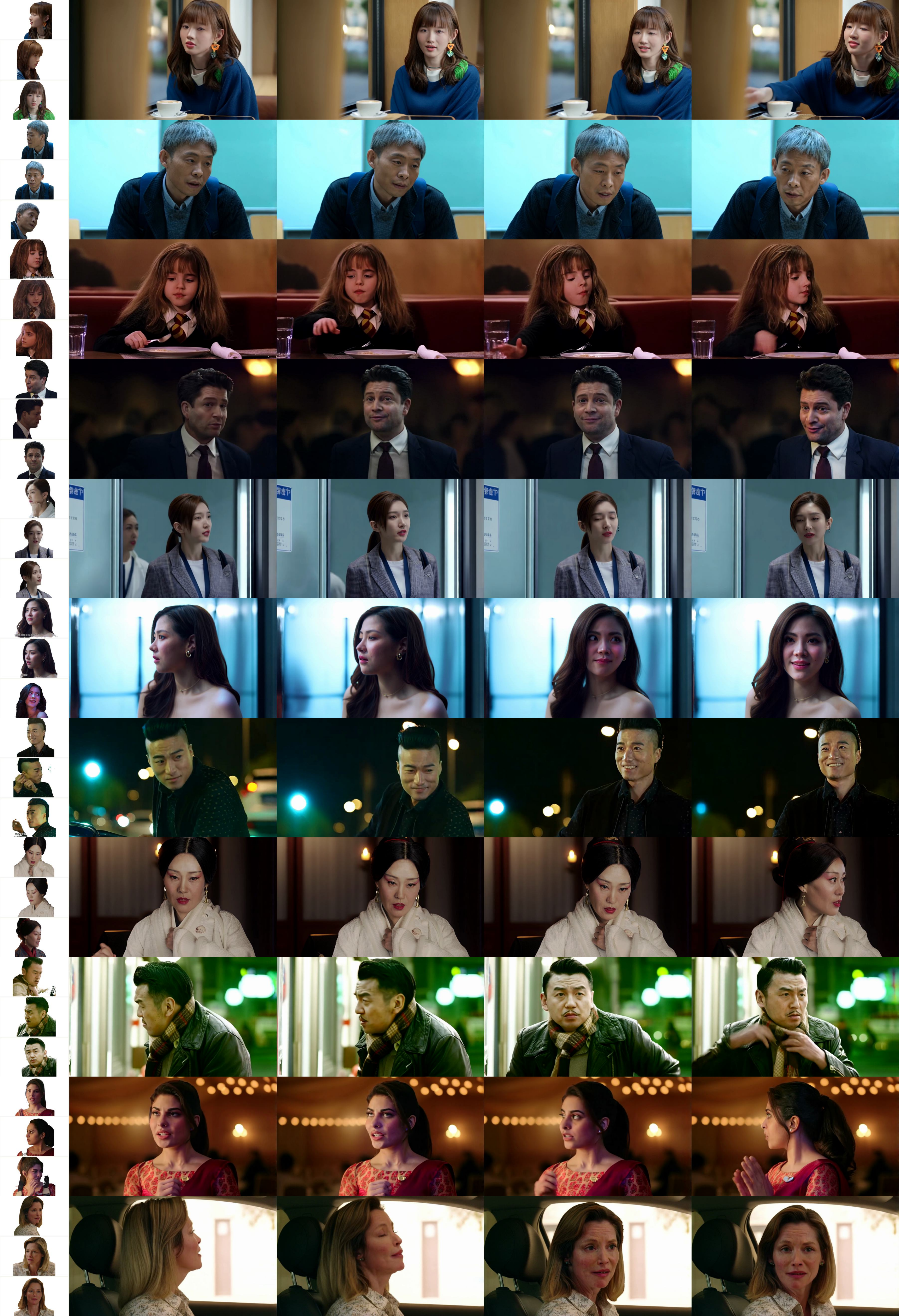}
    \caption{More visual results of our method.}
    \label{fig:more-visual}
\end{figure}

We provide additional visual results of our method in Fig.~\ref{fig:more-visual}. We additionally try some widely used strategies, such as using dynamic number of reference images and shuffling range of reference images. These strategies can significantly improve the quality of results. However, in order to avoid confusing the analysis of the proposed RM and RD-RoPE, we do not include the these results in the main text.

\subsection{More Facial Trajectory Results}
\begin{figure}[t]
    \centering
    \includegraphics[width=0.95\linewidth]{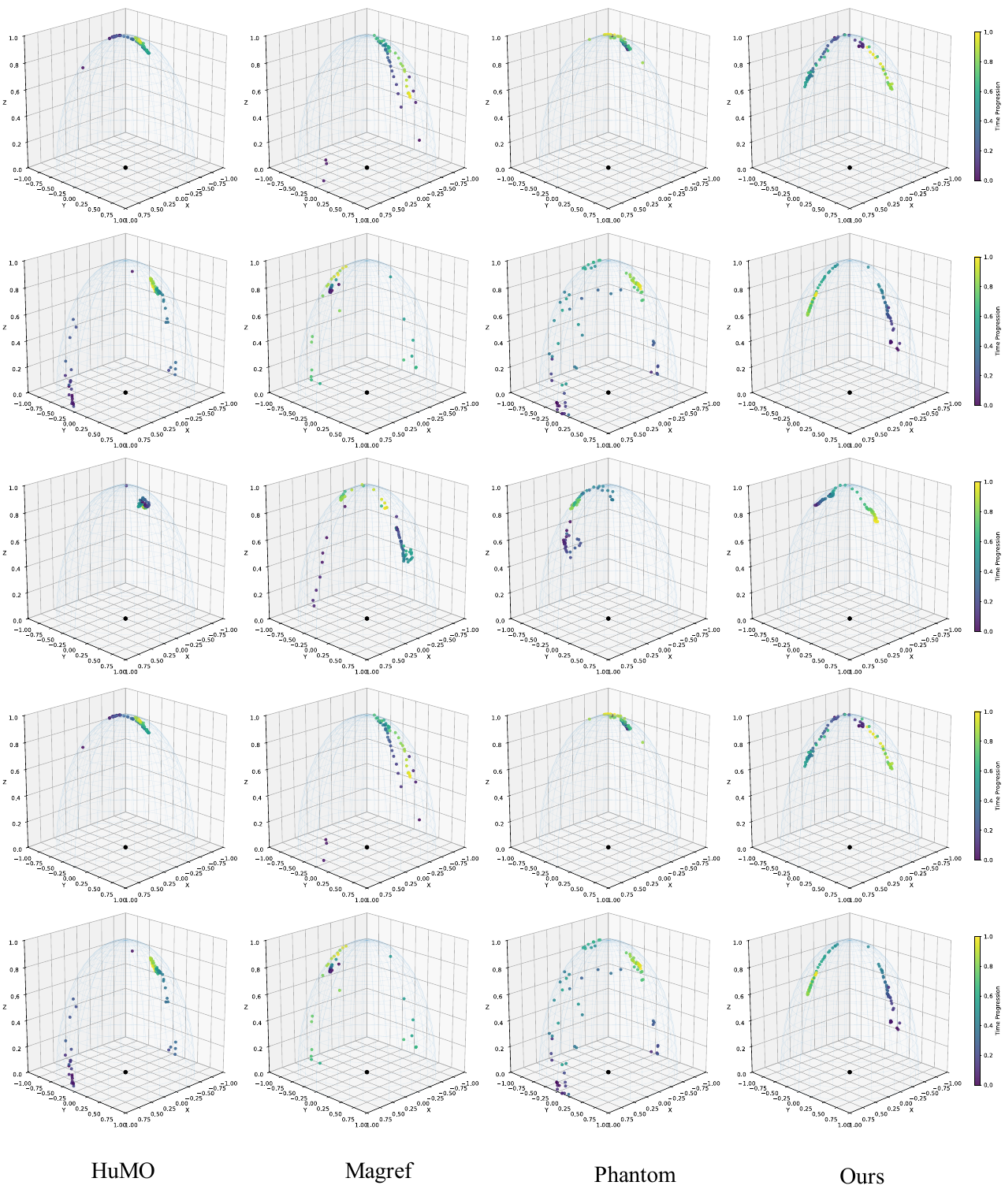}
    \caption{The facial trajectory results of our method and compared methods.}
    \label{fig:more-trajectory}
\end{figure}

We provide more facial trajectory results in Fig.~\ref{fig:more-trajectory} of our method and baseline methods. From the results, we can clearly find that the facial trajectories of our method are more smooth and evenly distributed, indicating more natural facial motion.


%
%

\end{document}